\documentclass{article}

    \PassOptionsToPackage{numbers, compress}{natbib}
 \usepackage[preprint]{neurips_2026}

\usepackage[utf8]{inputenc} 
\usepackage[T1]{fontenc}    
\usepackage{hyperref}       
\usepackage{url}            
\usepackage{booktabs}       
\usepackage{amsfonts}       
\usepackage{nicefrac}       
\usepackage{microtype}      
\usepackage{xcolor}         
\usepackage{todonotes}
\usepackage{xspace}
\usepackage{amsmath}
\usepackage{mwe}
\usepackage{float}
\usepackage{multirow}
\usepackage{enumitem}
\usepackage{comment}
\usepackage{subcaption}
\usepackage{amssymb}       

\newif\ifshowrev
\showrevfalse
\newcommand{\rev}[1]{\ifshowrev{\color{blue}#1}\else{#1}\fi}
\newcommand{\revcolor}{\ifshowrev\color{blue}\fi} 

\newcommand{\methodname}{\textbf{GenRec}\xspace}

\title{GenRec: Knowing Where to Reconstruct and Where to Generate}

\author{%
  Ata \c{C}elen$^{1}$ \quad
  Jaewoo Jung$^{1,4}$ \quad
  Federico Tombari$^{2}$ \quad
  Marc Pollefeys$^{1,3}$ \\[2pt]
  \textbf{Sunghwan Hong$^{1}$ \quad
  Michael Niemeyer$^{2}$ \quad
  Daniel Barath$^{1,2}$} \\[6pt]
  $^{1}$ETH Z\"urich \quad
  $^{2}$Google \quad
  $^{3}$Microsoft \quad
  $^{4}$KAIST
}

\begin{document}

\maketitle

\begin{center}
\large\href{https://atcelen.github.io/GenRec/}{\texttt{https://atcelen.github.io/GenRec/}}
\end{center}

\begin{abstract}
Generative novel view synthesis from sparse input images is rarely all reconstruction or all generation: pixels visible in some source view have a unique correct value modulated only by view-dependent shading, while pixels in disocclusions or beyond the captured volume admit a distribution of plausible completions. Existing generative novel-view-synthesis methods conflate these regimes under a single uniform loss, blurring the line between geometric fidelity and creative hallucinations even when scene geometry is injected through warped point clouds or projected depth. We introduce GenRec, a multi-view flow matching model that builds the reconstruction--generation split directly into its architecture, supervision, and gradient flow. Guided by an observation mask derived from the source cameras and a monocular depth estimator, a flow matching backbone jointly denoises RGB and scene-coordinate maps across all target views, while a pixel-space refinement stage restores high-frequency detail on observed pixels; the same mask gates supervision so regression signals do not contaminate the generative prior. Across RealEstate10K, DL3DV-10K, and Mip-NeRF~360, in both single-view extrapolation and two-view interpolation, GenRec attains the best reconstruction fidelity in observed regions while also surpassing purely generative baselines on perceptual quality in unobserved ones, showing the effectiveness of our approach.
\end{abstract}

\section{Introduction}

Consider two scenarios: a robot navigates an indoor space from a handful of casually captured photographs, while a headset user wanders freely through the same scene in mixed reality. The robot must plan collision-free trajectories around furniture seen from a single angle, judge whether a gap between two chairs is wide enough to traverse, and anticipate what lies beyond a half-open door before committing to a turn. The headset user expects every captured chair and painting to remain exactly in its place from any angle, and any corridor they step into beyond what was captured to fill in seamlessly around them. All of these capabilities reduce to a single problem: generative novel-view synthesis with sparse input views. Yet within a single synthesized view, different pixels carry very different burdens. Where the new view overlaps with what the camera already observed, the system must \emph{reconstruct} the scene faithfully: the same wall, the same texture, modulated only by view-dependent effects (e.g., specular highlights) arising from the angular shift. Where the new view ventures into unobserved regions (around a corner, behind an occluder, or beyond the captured volume), the inputs alone do not pin down a single answer, and the system must \emph{infer} the most likely completion consistent with the visible evidence and a learned prior over the world.


Today's 3D scene representations, such as 3D Gaussian Splatting (3DGS)~\citep{kerbl20233d} and NeRFs~\citep{mildenhall2021nerf}, only handle the first half. They are interpolators: dense observations yield photorealistic in-between views, but they have nothing to say about the parts of the world the camera never looked at, and degrade sharply as input becomes sparse~\citep{warburg2023nerfbusters, niemeyer2022regnerf}. To bridge this gap, a recent line of \emph{generative reconstruction} methods leverages image and video diffusion priors~\citep{wu2024reconfusion, gao2024cat3d, yu2024viewcrafter, ren2025gen3c, zhou2025stable, liang2025wonderland, szymanowicz2025bolt3d, bahmani2025lyra, liu2026reconx} to hallucinate plausible novel views from one or a few input views, which can then be fed back as pseudo-observations to a 3DGS or NeRF~\citep{wu2025difix3d+}. These methods differ mainly in how they tell the generator \emph{where} to look: pose-only conditioning via Pl\"ucker rays~\citep{he2024cameractrl, gao2024cat3d, zhou2025stable, jin2024lvsm} suffers from metric-scale ambiguity, while warp-based conditioning that injects forward-projected depth or point-cloud renders~\citep{yu2024viewcrafter, ren2025gen3c, fischer2025flowr} grounds the model in the scene but inherits warp artifacts.

For all these methods, the generator must produce \emph{every} pixel under a single distribution-matching loss. Yet, this overlooks a basic structural property of the task. A novel view is rarely all generation or all reconstruction; it is a combination of both. Pixels on a surface visible in the captured scene have a unique correct answer, dictated by geometry and appearance. Pixels in disocclusions or beyond the captured volume have many plausible answers, and the model must commit to a coherent one. Asking a single network to satisfy both regimes with the same supervisory signal forces a compromise: a loss that fits a distribution where many answers are valid penalizes fidelity where only one is. 

We propose \methodname, which incorporates this separation directly into its architecture: the network \emph{reconstructs} observed regions and \emph{infers} unobserved ones. Given the input cameras and an off-the-shelf depth estimator, we forward-project source pixels into each target view to obtain a binary observed/unobserved mask. The generative backbone produces a full target view; pixel-space refinement then operates on the observed pixels, leveraging warped input evidence and learned 3D correspondences to recover the high-frequency detail suppressed by latent flow matching. These modules do not share parameters with the backbone and their gradients do not propagate into it. Thus, the distribution-matching objective is no longer required to anchor pixels whose answer is uniquely determined by geometry; the refinement modules supply this fidelity signal on observed pixels alone.

Our contributions are:
%
%
\begin{itemize}[nosep]
    \item A \emph{geometry-grounded decomposition} of generative NVS that processes observed/unobserved pixels with separate modules and losses, governed by a per-pixel observation mask.
    \item A \emph{multi-view flow matching backbone} that jointly denoises RGB and scene-coordinate maps, yielding cross-view consistency and a 3D correspondence signal in a single pass.
    \item A \emph{pixel-space reconstruction branch} with sparse 3D cross-attention that recovers detail lost to latent compression without disturbing the generative prior in unobserved regions.
\end{itemize}

We validate \methodname{} on RealEstate10K, DL3DV-10K, and Mip-NeRF~360 under both single-view extrapolation and two-view interpolation. Across these settings, \methodname{} achieves the best reconstruction fidelity in observed regions while surpassing purely generative baselines on perceptual quality in unobserved ones, demonstrating that the separation of these two tasks yields both better reconstruction and generation quality. At the same time, our model is two orders of magnitude faster at inference than the strongest baseline.

\section{Related Work} 


\begin{figure}[t]
  \centering
  \begin{subfigure}{0.45\linewidth}
    \centering
    \includegraphics[width=\linewidth]{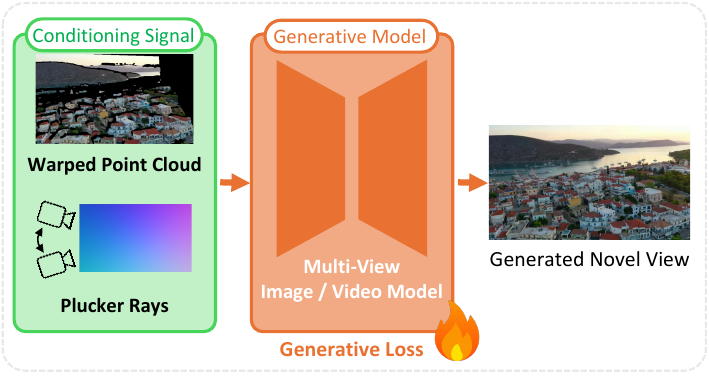}
    \caption{Prior generative reconstruction works}
    \label{fig:framework_prior}
  \end{subfigure}
  \hfill
  \begin{subfigure}{0.49\linewidth}
    \centering
    \includegraphics[width=\linewidth]{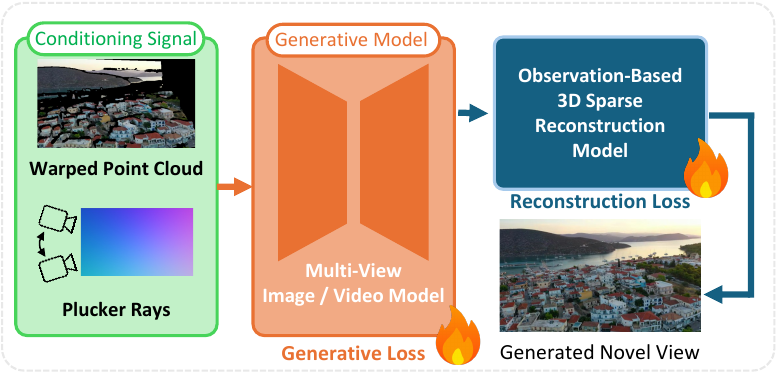}
    \caption{\methodname{}}
    \label{fig:framework_ours}
  \end{subfigure}
  \caption{\textbf{Overview of \methodname{} (right) vs.\ prior generative reconstruction methods (left).} Prior methods condition multi-view image or video models on warped point clouds or Pl\"ucker rays, supervising output pixels with one generative loss regardless of source coverage. \methodname{} factors the task by an observation mask: a generative backbone predicts every pixel, and a pixel-space reconstruction module refines the observed ones under a dedicated loss. The two pathways share neither parameters nor gradients, preserving the generative prior in unobserved regions.}
  \label{fig:framework}
\end{figure}

\vspace{1mm}\noindent\textbf{Novel view synthesis from sparse inputs.}
Neural Radiance Fields~\citep{mildenhall2021nerf} and 3D Gaussian Splatting~\citep{kerbl20233d}, together with their follow-ups for unbounded scenes~\citep{barron2022mip} and faster rendering~\citep{barron2023zip}, set the modern bar for in-distribution view synthesis. They are, however, fundamentally interpolators: they require dense captures, are optimized per scene, and degrade once the camera leaves the convex hull of the inputs~\citep{warburg2023nerfbusters}. Geometric priors such as depth smoothness~\citep{niemeyer2022regnerf} or monocular cues~\citep{yu2022monosdf} partially mitigate the sparse-view regime by regularizing the optimization. Feed-forward alternatives sidestep per-scene fitting altogether: PixelNeRF~\citep{yu2021pixelnerf} conditions a radiance field on image features in a single forward pass, and recent work predicts 3D Gaussians directly from sparse views~\citep{charatan2024pixelsplat, chen2024mvsplat, ye2024no, tang2024lgm, xu2025depthsplat, ye2025yonosplat}, while large transformer synthesizers such as LVSM~\citep{jin2024lvsm} and projection-conditioned variants~\citep{wu2026rays} dispense with explicit 3D representations entirely. Across this entire family of methods, the output is deterministic, which is sufficient when the target view is well-constrained but breaks down when large regions must be imagined; the gap that the next family fills.

\vspace{1mm}\noindent\textbf{Pose-only multi-view diffusion.}
These methods condition a multi-view diffusion model purely on the target camera, typically through Pl\"ucker ray embeddings~\citep{sitzmann2021light, he2024cameractrl}. Beginning with score distillation~\citep{poole2022dreamfusion} and image-conditioned single-object generators~\citep{liu2023zero, shi2023zero123++, shi2023mvdream}, it has scaled to scene-level diffusion that produces tens of consistent novel views in a single denoising pass~\citep{wu2024reconfusion, gao2024cat3d, zhou2025stable}, and to joint RGB-plus-pointmap sampling for fast 3D scene generation~\citep{szymanowicz2025bolt3d}. Pose-only conditioning is flexible but struggles with the metric-scale ambiguity of ray embeddings. Crucially, the content of the scene actually visible in the input is not given any privileges: it is regenerated from scratch like the rest of the frame.

\vspace{1mm}\noindent\textbf{Warp- and geometry-conditioned generation.}
Another line of work explicitly injects scene geometry into the diffusion model. The most common recipe takes a depth-warped point cloud or RGB-D rendering of the source views and feeds it as a frame-aligned signal to a video diffusion backbone~\citep{yu2024viewcrafter, liang2025wonderland, bahmani2025lyra, liu2026reconx}. GEN3C~\citep{ren2025gen3c} formalizes this with a persistent 3D cache that is rendered into the model for precise camera control, and Geometry Forcing~\citep{wu2025geometry} pushes the video backbone toward 3D-consistent latents via auxiliary feature alignment. FlowR~\citep{fischer2025flowr} replaces video diffusion with multi-view flow matching that maps renderings of a sparse 3DGS to renderings of a dense one. A complementary thread bridges geometric foundation models with diffusion latents~\citep{wang2025vggt, lin2025depth, huang2026gen3r, jang2026repurposing}, and a third uses generative priors to clean or harmonize existing reconstructions~\citep{wu2025difix3d+}. These methods inject geometry where pose-only models cannot, but the supervision they apply remains uniform across the output: regions where the warped condition is reliable and regions where it is empty are pulled toward the same target by the same loss.

\vspace{1mm}\noindent\textbf{Region-aware supervision.}
Binary masks do appear in the geometry-coupled literature, but always at the periphery of the generative loop: dropping parts of the 3D representation during training~\citep{wu2025genfusion}, marking unwarped pixels in the input conditioning~\citep{yu2024viewcrafter}, or reweighting rendering losses in the downstream 3DGS optimization. None of these methods uses a per-pixel observation signal to split reconstruction from generation inside the generator itself. A related observation, recently sharpened by pixel-space NVS diffusion~\citep{elata2025novel}, is that VAE latents impose a photometric fidelity ceiling on the detail that observed regions can carry through. \methodname{} addresses both: a shared observation mask binds architecture and loss, and a dedicated pixel-space branch restores detail where the inputs constrain it while leaving the generative prior untouched elsewhere.

\section{Method}
\label{sec:method}

\begin{figure}[t]
  \centering
  \includegraphics[width=\linewidth]{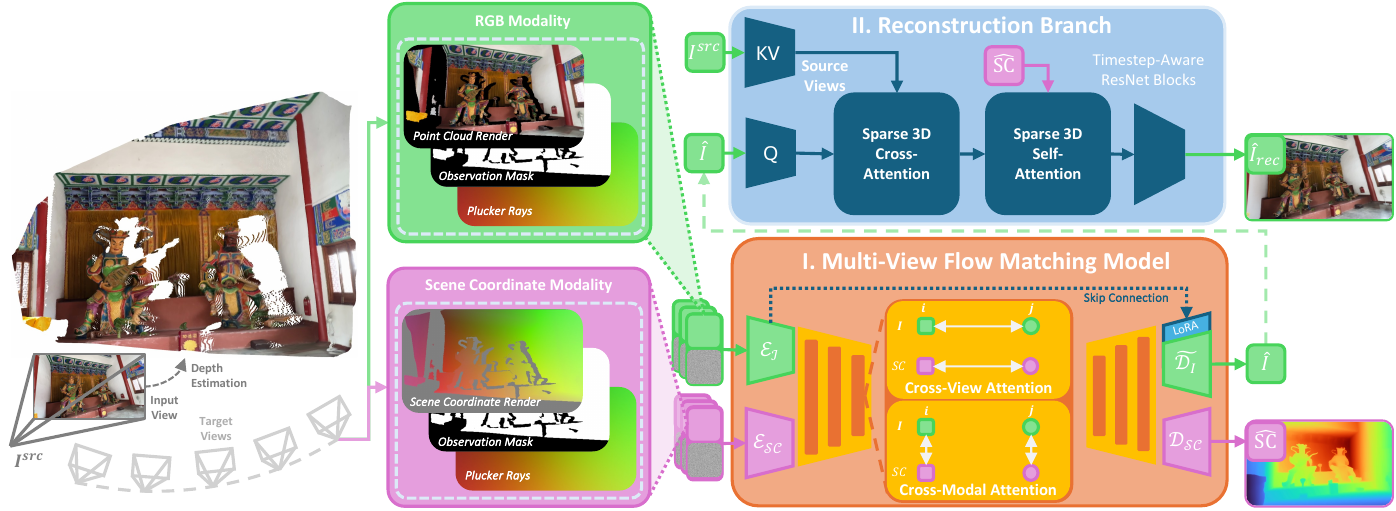}
  \caption{\textbf{Overview of \methodname{}}. Given posed input views and target poses, monocular depth and forward warping produce per-target observation masks and warped renders. These condition a \textit{(I) multi-view flow matching backbone}, which jointly denoises RGB and scene-coordinate latents through cross-view and cross-modal attention. A \textit{(II) reconstruction branch} then refines only the observed pixels via sparse 3D attention guided by the predicted scene coordinates, while the observation mask gates gradients so the generative prior is preserved in unobserved regions.}
  \label{fig:pipeline}
\end{figure}

\noindent\textbf{Problem statement.}
We are given $N_{\text{src}}$ posed source views $\{(I^{\text{src}}_j,\pi^{\text{src}}_j)\}_{j=1}^{N_{\text{src}}}$ with images $I^{\text{src}}_j\in[0,1]^{3\times H\times W}$, and $N$ target poses $\{\pi_i\}_{i=1}^{N}$. The task is to synthesize the corresponding target images $\{\hat{I}_i^{\,\mathrm{rec}}\}_{i=1}^{N}$, with $\hat{I}_i^{\,\mathrm{rec}}\in[0,1]^{3\times H\times W}$. As preprocessing, we run an off-the-shelf monocular depth prediction network on each source view and forward-warp the source RGB and the back-projected source 3D points into every target frame. This yields, for each target view $i$, three precomputed signals: a soft observation mask $O_i\in[0,1]^{1\times H\times W}$ recording which target pixels receive evidence, and per-modality warped renderings $W_i^{\,I}\in[0,1]^{3\times H\times W}$ of the source RGB and $W_i^{\,SC}\in[-1,1]^{3\times H\times W}$ of the source scene coordinates.


\noindent\textbf{Generation--reconstruction split.}
Not all pixels in a target view are equally constrained by the source views. Pixels with source coverage have a near-deterministic target: the source observation collapses the predictive distribution to a tight mode varying only with viewing direction. Pixels without coverage (in disocclusions or beyond the captured frustum) have a stochastic target: only a posterior over completions consistent with the visible scene is identifiable. These regimes demand structurally different supervision (regression in the first, distribution matching in the second), and the latent generative path discards the high-frequency content the deterministic regime must preserve. \methodname{} therefore factors the model along $O_i$ at every level (Fig.~\ref{fig:pipeline}). A multi-view flow-matching backbone (Sec.~\ref{sec:backbone}) generates every pixel of every target view, handling the distributional regime. A lightweight pixel-space reconstruction stage (Sec.~\ref{sec:refine}) then corrects the observed pixels, supplying the photometric signal the latent path cannot provide. The mask further isolates gradient flow: reconstruction gradients reach only the reconstruction parameters, never the backbone (Sec.~\ref{sec:training}).

\subsection{Multi-View Flow Matching Backbone}
\label{sec:backbone}

The backbone is a denoising network $v_\theta$ trained under the Conditional Flow Matching framework. It transports Gaussian noise $x_0\sim\mathcal{N}(0,I)$ to a clean latent $x_1$ along the path $x_t=(1-t)\,x_0+t\,x_1$. A single denoising trajectory operates jointly on all $N$ target views and two co-registered modalities, producing per-view, per-modality clean-end estimates $\hat{x}_{1,i}^{\,m}\in\mathbb{R}^{C\times h\times w}$ for $i\in\{1,\dots,N\}$ and $m\in\{I,SC\}$. The refinement stage of Sec.~\ref{sec:refine} consumes these latents.

\noindent\textbf{Joint RGB and scene-coordinate denoising.}
Each target view comprises two co-registered modalities: the RGB image and a scene-coordinate map $SC\in[-1,1]^{3\times H\times W}$ encoding the normalized world-space $(x,y,z)$ position of every pixel. Joint denoising enables the geometry-aware refinement of Sec.~\ref{sec:refine}: the decoded scene coordinates $\widehat{SC}_i=\mathcal{D}_{SC}(\hat{x}_{1,i}^{\,SC})$ supply the 3D positions used by its cross-attention. Both modalities use pretrained VAEs $(\mathcal{E}_I,\mathcal{D}_I)$ and $(\mathcal{E}_{SC},\mathcal{D}_{SC})$ with shared channel count $C$ and spatial resolution $h\times w$. Following~\citet{fang2025spatialgen}, we extend each transformer block of $v_\theta$ along two axes: (i) self-attention is rewired into \emph{cross-view} attention over all $N$ target tokens, yielding multi-view consistency in one forward pass; (ii) a zero-initialized \emph{cross-modal} attention layer couples the RGB and scene-coordinate tokens of the same view. The zero initialization preserves the pretrained prior at training start; the modalities are paired as their joint distribution is learned.

\noindent\textbf{Conditioning on geometric evidence.}
At each flow time $t\in[0,1]$, the network input is a tensor $\mathbf{Z}_{\mathrm{in}}(t)\in\mathbb{R}^{N\times 2\times C_{\mathrm{in}}\times h\times w}$ where the second axis indexes the two modalities and $C_{\mathrm{in}}$ is the per-modality concatenated channel count. For every view $i$ and modality $m\in\{I,SC\}$,
\begin{equation}
    \mathbf{Z}_{\mathrm{in}}^{\,i,m}(t) \;=\; \big[\,x_{t,i}^{\,m}\,\big\Vert\,\mathbf{r}_i\,\big\Vert\,\tilde{O}_i\,\big\Vert\,\mathcal{E}_m\!\big(W_i^{\,m}\big)\,\big],
    \label{eq:conditioning}
\end{equation}
where $\Vert$ denotes channel-wise concatenation; $x_{t,i}^{\,m}$ is the noisy latent at time $t$; $\mathbf{r}_i$ is a Pl\"ucker ray map of the target, embedded by a small patch ViT~\citep{dosovitskiy2020image} whose tokens are folded back to the latent grid; $\tilde{O}_i$ is the observation mask area-pooled to $h\times w$; and $\mathcal{E}_m(W_i^{\,m})$ encodes the modality-specific warp. This tells the backbone \emph{where} evidence is available (via $\tilde{O}_i$), \emph{what} that evidence is (via the encoded warps), and \emph{which} 3D ray each pixel corresponds to (via $\mathbf{r}_i$), even in regions it must hallucinate.

\noindent\textbf{Flow-matching parameterization.}
We initialize $v_\theta$ from a pretrained $\mathbf{v}$-prediction latent diffusion model~\citep{rombach2022high} and fine-tune it under Conditional Flow Matching with the Diff2Flow reparameterization of~\citet{schusterbauer2025diff2flow}, which converts the $\mathbf{v}$-prediction into a flow velocity in closed form. This preserves the pretrained weights (cross-modal layer is zero-initialized) while letting us train straight flows that admit fast Euler sampling at inference~\citep{liu2022flow}. The clean estimate is recovered in closed form as $\hat{x}_{1,i}^{\,m}=x_{t,i}^{\,m}+(1-t)\,v_\theta(\mathbf{Z}_{\mathrm{in}}(t),t)_{i,m}$, where $(\cdot)_{i,m}$ selects the per-view, per-modality slice.

\subsection{Pixel-Space Refinement of Observed Regions}
\label{sec:refine}

The refinement stage takes the backbone's clean latents $\{\hat{x}_{1,i}^{\,I},\hat{x}_{1,i}^{\,SC}\}_{i=1}^{N}$ together with the precomputed mask $O_i$ and warps $W_i^{\,I}$, and produces the final RGB outputs $\{\hat{I}_i^{\,\mathrm{rec}}\}_{i=1}^{N}$. Latent generative models impose a photometric ceiling on what can be recovered through the VAE decoder (acceptable for hallucinated content, but limiting where evidence already constrains the answer), so we restrict refinement to observed pixels. The stage has two cooperating components: a geometry-aware decoder adapter that injects warped RGB at the latent$\to$pixel boundary and outputs an intermediate image $\hat{I}_i$, and a sparse 3D cross-attention branch that adds an explicitly geometry-aware residual $\Delta_i$ on top.

\noindent\textbf{Geometry-Aware Decoder Adapter.}
We adapt the RGB decoder $\mathcal{D}_I$, which maps the latent $\hat{x}_{1,i}^{\,I}$ to pixels; both VAEs are otherwise frozen. The adaptation has two parts. Following~\citet{parmar2024one,wu2025difix3d+}, zero-initialized $1{\times}1$ skip convolutions additively inject features from $\mathcal{E}_I(W_i^{\,I})$ before each up-block of $\mathcal{D}_I$. A LoRA adapter~\citep{hu2022lora} on the decoder's convolutions and self-attention projections is jointly trained, allowing the decoder to consume these injected features. We denote the resulting decoder $\widetilde{\mathcal{D}}_I$, with output:
\begin{equation*}
    \hat{I}_i \;=\; \widetilde{\mathcal{D}}_I\!\big(\hat{x}_{1,i}^{\,I},\,W_i^{\,I}\big) \;\in\;[0,1]^{3\times H\times W}.
\end{equation*}
The two adaptations are complementary. \emph{Locally}, the skip connections copy fine-grained photometric detail from the warped conditioning, recovering high-frequency content the latent representation cannot preserve. \emph{Globally}, the LoRA weights affect every pixel and shift the decoder's output distribution toward sharper samples, so perceptual benefits extend beyond the observation mask. Supervision is gated by $O_i$ (Sec.~\ref{sec:training}) to prevent drift in disoccluded regions.

\noindent\textbf{Sparse 3D Reconstruction Branch.}
The reconstruction branch consumes the decoder-adapter output $\hat{I}_i$ together with the warped RGB $W_i^{\,I}$, the mask $O_i$, the source images $\{I^{\text{src}}_j\}_{j=1}^{N_{\text{src}}}$, and the decoded scene coordinates $\widehat{SC}_i=\mathcal{D}_{SC}(\hat{x}_{1,i}^{\,SC})$. It predicts a pixel-space residual $\Delta_i\in\mathbb{R}^{3\times H\times W}$ that is added back only on observed pixels, producing the final image
\begin{equation*}
    \hat{I}_i^{\,\mathrm{rec}} \;=\; \hat{I}_i \;+\; O_i\odot\Delta_i.
\end{equation*}
Whereas the decoder adapter copies coarse appearance through skip connections, this branch reasons explicitly about 3D correspondences: each target pixel attends to a small set of geometrically nearest neighbours (first, in the source views, then across target views), with $\widehat{SC}_i$ supplying the 3D positions used to define the neighbour sets. The branch is composed of two sparse cross-attention blocks followed by FiLM-conditioned residual blocks~\citep{perez2018film} whose scale and shift are produced from the detached temporal embedding of the backbone, so the branch modulates its behaviour with the flow time $t$. The output projection of every cross-attention block and final convolution are zero-initialized, so $\Delta_i=\mathbf{0}$ at the start of training and the branch contributes only as it acquires useful signal.

\noindent\textit{Sparse source-to-target attention.}
The first attention block operates at full pixel resolution. For each target view $i$, queries and values are built from the per-view tensor $[\hat{I}_i\,\Vert\,W_i^{\,I}\,\Vert\,O_i]$, and keys are produced by encoding the source images $\{I^{\text{src}}_j\}$ with a VAE-initialized encoder. To avoid quadratic cost across all source pixels, we precompute, for each target pixel $n$, the indices of its $K$ geometrically nearest source pixels using $\widehat{SC}_i$ and the source point cloud, and restrict attention to that fixed neighbour set. Indexing those neighbours by $k\in\{1,\dots,K\}$, we form the attention logits:
\begin{equation}
    \mathrm{attn}_{n,k}
    \;=\;
    \frac{Q_n^{\!\top} K_{n,k}}{\sqrt{d}}
    \;+\;
    \log w_{n,k},
    \qquad
    w_{n,k}
    \;=\;
    \frac{(d_{n,k}+\varepsilon)^{-1}}{\sum_{k'}(d_{n,k'}+\varepsilon)^{-1}},
    \label{eq:sparse_cross_attn}
\end{equation}
where $d$ is the head dimension, $\mathbf{p}_n^{\,\mathrm{tgt}}\in\mathbb{R}^3$ is the target-pixel 3D position read from $\widehat{SC}_i$, $\mathbf{p}_k^{\,\mathrm{src}}\in\mathbb{R}^3$ is the 3D position of its $k$-th source neighbour, and $d_{n,k}=\|\mathbf{p}_n^{\,\mathrm{tgt}}-\mathbf{p}_k^{\,\mathrm{src}}\|_2$. Since $\sum_k w_{n,k}=1$, the additive $\log w_{n,k}$ acts as a soft logit prior favouring geometrically close correspondences while still allowing evidence to override the prior; invalid slots are masked to $-\infty$ before the softmax.

\noindent\textit{Sparse cross-target attention.}
Per-view source-to-target attention may produce refinements that disagree where different target frames observe the same surface. A second attention block re-imposes consistency by reusing the same operator, but with the neighbour set now constructed across target views: for each target pixel $n$ in view $i$, we use $\widehat{SC}_i$ and $\widehat{SC}_{i'\ne i}$ to retrieve the $K$ geometrically nearest pixels in the \emph{other} target views. Queries come from the residual stream of the source-to-target stage, and keys and values are produced by re-encoding the per-view tensor $[\hat{I}_{i'}\,\Vert\,W_{i'}^{\,I}\,\Vert\,O_{i'}]$ from each other view. Eq.~\eqref{eq:sparse_cross_attn} is reused unchanged, with $d_{n,k}$ now measuring inter-target distance.

\subsection{Training and Inference}
\label{sec:training}

\noindent\textbf{Decoupled optimization.}
We train the model in two phases. In the first phase, only the multi-view flow-matching backbone is optimized, using the standard Conditional Flow Matching loss of Eq.~\eqref{eq:cfm_loss}, denoted $\mathcal{L}_{\mathrm{CFM}}$. In the second phase, the backbone is frozen and we train the decoder adapter and the reconstruction branch on its detached outputs, so no gradient from this phase can reach the backbone parameters. To prevent the adapters from drifting in disoccluded regions, gradients through the global LPIPS loss~\citep{zhang2018unreasonable} are routed only through observed pixels using the masked stop-gradient construction:
\begin{equation}
    \tilde{Y}_i \;=\; O_i\odot Y_i \;+\; (1-O_i)\odot \mathrm{sg}(Y_i),
    \label{eq:masked_sg}
\end{equation}
applied to each output of interest $Y_i\in\{\hat{I}_i,\hat{I}_i^{\,\mathrm{rec}}\}$.

\noindent\textbf{Refinement losses.}
The decoder-adapter output is supervised with a masked $\ell_1$ photometric term and a full-image LPIPS term gated by Eq.~\eqref{eq:masked_sg},
\begin{equation}
    \mathcal{L}_{\mathrm{dec}}
    \;=\;
    \sum_{i=1}^{N}\!\Big[\;
    \lambda_{1}\,\big\|O_i\odot(\hat{I}_i-I_i)\big\|_1
    \;+\;
    \lambda_{\mathrm{p}}\,\mathcal{L}_{\mathrm{LPIPS}}\!\big(\tilde{I}_i,\,I_i\big)
    \;\Big],
    \label{eq:decoder_loss}
\end{equation}
and the refined output by an analogous loss with an additional $\ell_1$ regularizer on $\Delta_i$ that keeps the refinement small by default,
\begin{equation}
    \mathcal{L}_{\mathrm{rec}}
    \;=\;
    \sum_{i=1}^{N}\!\Big[\;
    \lambda_{1}\,\big\|O_i\odot(\hat{I}_i^{\,\mathrm{rec}}-I_i)\big\|_1
    \;+\;
    \lambda_{\mathrm{p}}\,\mathcal{L}_{\mathrm{LPIPS}}\!\big(\tilde{I}_i^{\,\mathrm{rec}},\,I_i\big)
    \;+\;
    \lambda_{\Delta}\,\big\|\Delta_i\big\|_1
    \;\Big].
    \label{eq:recon_loss}
\end{equation}
The second phase optimizes $\mathcal{L}_{\mathrm{dec}}+\mathcal{L}_{\mathrm{rec}}$ over the decoder adapter and reconstruction branch with the backbone kept fixed. Combined with the $O_i$-gating in Eq.~\eqref{eq:masked_sg}, this two-phase schedule implements the generation--reconstruction split at the level of every gradient that touches the network.

\noindent\textbf{Iterative-denoising supervision for the refinement stage.}
A subtle but important design choice concerns \emph{which} clean-end estimate $\hat{x}_{1,i}^{\,I}$ is used to train the refinement stage. The naive recipe samples $t\in[0,1]$, forms the linear interpolant $x_{t,i}^{\,I}=(1-t)\,x_{0,i}^{\,I}+t\,x_{1,i}^{\,I}$, and feeds the corresponding one-step estimate $\hat{x}_{1,i}^{\,I}=x_{t,i}^{\,I}+(1-t)\,v_\theta(\mathbf{Z}_{\mathrm{in}}(t),t)_{i,I}$ to the refinement stage. This is misaligned with test-time behaviour: at inference, the refinement stage operates on a $\hat{x}_{1,i}^{\,I}$ produced from a multi-step Euler trajectory that has accumulated errors, whereas $x_{t,i}^{\,I}$ lies exactly on the noise–data line. We instead sample $t\in[0,1]$, integrate the flow from pure noise up to time $t$ to obtain a trajectory state $\hat{x}_{t,i}^{\,I}$, and supervise the refinement stage on $\hat{x}_{1,i}^{\,I}=\hat{x}_{t,i}^{\,I}+(1-t)\,v_\theta(\mathbf{Z}_{\mathrm{in}}(t),t)_{i,I}$. 

\noindent\textbf{Inference.}
Given source views and target poses, we precompute $O_i$, $W_i^{\,I}$, $W_i^{\,SC}$, and the geometric neighbour indices. Starting from $x_{0,i}^{\,m}\sim\mathcal{N}(0,I)$, we run $T$ Euler steps of the joint RGB+SC flow in a single multi-view denoising trajectory, yielding $\hat{x}_{1,i}^{\,I}$ and $\hat{x}_{1,i}^{\,SC}$. Decoding $\hat{x}_{1,i}^{\,I}$ through $\widetilde{\mathcal{D}}_I$ produces $\hat{I}_i$; a single forward pass through the reconstruction branch then yields $\Delta_i$ and the final $\hat{I}_i^{\,\mathrm{rec}}$. Both stages share the observation mask $O_i$, so the generation--reconstruction split holds at test time as well.

\vspace{-2mm}
\section{Experiments}
\label{sec:experiments}
\vspace{-2mm}

We organize our experiments into four groups. First, we evaluate \methodname{}'s extrapolation capabilities in the single-view conditioned setting. Second, we assess its interpolation abilities in the two-view conditioned setting, where target views lie between the two source views. Third, we measure the multi-view consistency of the generated frames using 3D-based metrics. Finally, we conduct ablation studies that validate our design choices and quantify the impact of our observation-based formulation on overall performance.



\noindent\textbf{Experimental Setup.}
We evaluate in two settings. \emph{Single-view extrapolation} uses the first frame of each test sequence as the source view and samples target views from subsequent frames at a fixed stride. \emph{Two-view interpolation} additionally uses the last frame as a second source. The stride is $4$ on DL3DV-10K and Mip-NeRF 360, and $25$ on RE10K to account for its slower camera motion. Video-architecture baselines receive the full pose sequence and produce outputs at the strided frames. Baselines conditioned on forward-warped RGB renders use DA3 for depth estimation and pose-depth alignment, matching \methodname{}. All baselines are open-source; we follow the evaluation protocols from their official repositories.



\noindent\textbf{Single-View Conditioned Extrapolation.}
We evaluate on two in-distribution benchmarks, RealEstate10K~\citep{zhou2018stereo} (RE10K) and DL3DV-10K~\citep{ling2024dl3dv}, and on Mip-NeRF 360~\citep{barron2022mip} for out-of-distribution generalization. We sample 100 scenes from the official test splits of RE10K and DL3DV-10K and use all 9 Mip-NeRF 360 scenes. We report PSNR, SSIM, and LPIPS for reconstruction quality and FID for generation quality. Unlike prior work, we report reconstruction metrics globally and separately for observed and unobserved regions. For fair comparison, observed and unobserved regions are scored against a shared binary observation mask from Gen3C's forward-warping, applied to all methods. We compare against baselines spanning different conditioning strategies and architectures: CameraCtrl~\citep{he2024cameractrl}, ViewCrafter~\citep{yu2024viewcrafter}, Stable Virtual Camera (SEVA)~\citep{zhou2025stable}, GLD~\citep{jang2026repurposing}, Gen3R~\citep{huang2026gen3r}, and Gen3C~\citep{ren2025gen3c}. On RE10K, we additionally compare against Geometry Forcing (GF)~\citep{wu2025geometry} and LVSM~\citep{jin2024lvsm}, both RE10K-only methods.

\begin{table}[t]
  \caption{\textbf{Quantitative results on RealEstate10K} in the single-view extrapolation setting. \emph{Global} metrics are computed over the full target frame; \emph{Observed} restricts them to pixels that receive source evidence under the shared observation mask, and \emph{Unobserved} to pixels that do not. Inference time is reported per scene. Best results in \textbf{bold}.}
  \label{tab:results-re10k}
  \centering
  \small
  \setlength{\tabcolsep}{3.5pt}
  \begin{tabular}{l cccc cc cc | r}
    \toprule
    & \multicolumn{4}{c}{Global} & \multicolumn{2}{c}{Observed} & \multicolumn{2}{c}{Unobserved} & \multicolumn{1}{c}{}\\ 
    \cmidrule(lr){2-5} \cmidrule(lr){6-7} \cmidrule(lr){8-9}
    Method & PSNR $\uparrow$ & SSIM $\uparrow$ & LPIPS $\downarrow$ & FID $\downarrow$ & PSNR $\uparrow$ & SSIM $\uparrow$ & PSNR $\uparrow$ & SSIM $\uparrow$ & Inf. Time \\
    \midrule
    CameraCtrl~\cite{he2024cameractrl} & 13.40 & 0.4424 & 0.4761 & 57.42 & 15.53 & 0.4568 & 12.40 & 0.3935 & $\sim$12s \\
    ViewCrafter~\cite{yu2024viewcrafter} & 13.29 & 0.4792 & 0.4293 & 46.27 & 15.93 & 0.5196 & 11.83 & 0.3937 & $\sim$200s \\
    SEVA~\cite{zhou2025stable} & 13.50 & 0.4839 & 0.4674 & 41.64 & 15.97 & 0.4999 & 12.40 & 0.4189 & $\sim$20s \\
    GF~\cite{wu2025geometry} & 13.11 & 0.3806 & 0.4945 & 54.58 & 15.56 & 0.3986 & 11.96 & 0.3125 & $\sim$40s \\
    LVSM~\cite{jin2024lvsm} & 13.81 & 0.4094 & 0.5186 & 57.89 & 16.32 & 0.4224 & 12.61 & 0.3435 & $\sim$1s \\
    GLD~\cite{jang2026repurposing} & 13.73 & 0.4232 & 0.4699 & 44.29 & 16.07 & 0.4533 & 12.82 & 0.3765 & $\sim$42s \\
    Gen3R~\cite{huang2026gen3r} & 13.29 & 0.4747 & 0.5107 & 49.46 & 15.82 & 0.4862 & 12.14 & 0.4044 & $\sim$90s \\
    GEN3C~\cite{ren2025gen3c} & 15.24 & 0.5289 & 0.3624 & 35.68 & 18.13 & 0.5689 & 13.63 & 0.4529 & $\sim$1200s \\
    \textbf{Ours} & \rev{\textbf{17.05}} & \textbf{0.6071} & \textbf{0.3220} & \textbf{33.09} & \textbf{20.85} & \textbf{0.6738} & \rev{\textbf{14.18}} & \rev{\textbf{0.4743}} & \rev{$\sim$11s} \\
    \bottomrule
  \end{tabular}
\end{table}
\begin{table}[t]
  \caption{\textbf{Quantitative Results on DL3DV-10K.} We compare our method with previous works on single-view conditioned generation. We follow the same evaluation protocol as in Table \ref{tab:results-re10k}.}
  \label{tab:results-dl3dv}
  \centering
  \small
  \setlength{\tabcolsep}{3.5pt}
  \begin{tabular}{l cccc cc cc | r}
    \toprule
    & \multicolumn{4}{c}{Global} & \multicolumn{2}{c}{Observed} & \multicolumn{2}{c}{Unobserved} & \multicolumn{1}{c}{}\\ 
    \cmidrule(lr){2-5} \cmidrule(lr){6-7} \cmidrule(lr){8-9}
    Method & PSNR $\uparrow$ & SSIM $\uparrow$ & LPIPS $\downarrow$ & FID $\downarrow$ & PSNR $\uparrow$ & SSIM $\uparrow$ & PSNR $\uparrow$ & SSIM $\uparrow$ & Inf. Time \\
    \midrule
    CameraCtrl~\cite{he2024cameractrl} & 13.26 & 0.3346 & 0.4893 & 69.49 & 13.45 & 0.3418 & 12.76 & 0.2957 & $\sim$12s \\
    ViewCrafter~\cite{yu2024viewcrafter} & 17.57 & 0.5302 & 0.3092 & 46.36 & 18.75 & 0.5659 & 15.36 & 0.4282 & $\sim$200s \\
    SEVA~\cite{zhou2025stable} & 14.58 & 0.4300 & 0.4359 & 43.51 & 14.86 & 0.4401 & 13.92 & 0.3819 & $\sim$20s \\
    GLD~\cite{jang2026repurposing} & 15.54 & 0.4001 & 0.3979 & 51.85 & 16.10 & 0.4184 & 14.33 & 0.3368 & $\sim$42s \\
    Gen3R~\cite{huang2026gen3r} & 13.74 & 0.3754 & 0.5173 & 130.10 & 14.02 & 0.3886 & 12.95 & 0.3048 & $\sim$90s \\
    GEN3C~\cite{ren2025gen3c} & 18.47 & 0.5717 & 0.2860 & 44.54 & 19.66 & 0.6026 & 16.24 & 0.4802 & $\sim$1200s \\
    \textbf{Ours} & \textbf{19.80} & \textbf{0.6195} & \textbf{0.2188} & \textbf{32.36} & \textbf{21.35} & \textbf{0.6605} & \rev{\textbf{16.66}} & \rev{\textbf{0.4983}} & \rev{$\sim$11s} \\
    \bottomrule
  \end{tabular}
\end{table}
\begin{table}[t]
  \caption{\textbf{Quantitative results on Mip-NeRF 360.} Out-of-distribution comparison with prior methods in the single-view extrapolation setting. We follow the same evaluation protocol as in Table \ref{tab:results-re10k}.}
  \label{tab:results-mipnerf360}
  \centering
  \small
  \setlength{\tabcolsep}{3.5pt}
  \begin{tabular}{l cccc cc cc | r}
    \toprule
    & \multicolumn{4}{c}{Global} & \multicolumn{2}{c}{Observed} & \multicolumn{2}{c}{Unobserved} & \multicolumn{1}{c}{}\\ 
    \cmidrule(lr){2-5} \cmidrule(lr){6-7} \cmidrule(lr){8-9}
    Method & PSNR $\uparrow$ & SSIM $\uparrow$ & LPIPS $\downarrow$ & FID $\downarrow$ & PSNR $\uparrow$ & SSIM $\uparrow$ & PSNR $\uparrow$ & SSIM $\uparrow$ & Inf. Time \\
    \midrule
    CameraCtrl~\cite{he2024cameractrl} & 10.89 & 0.1870 & 0.6534 & 214.2 & 11.31 & 0.1869 & 10.67 & 0.1821 & $\sim$12s \\
    ViewCrafter~\cite{yu2024viewcrafter} & 11.75 & 0.2197 & 0.5694 & 172.5 & 14.46 & 0.2510 & 10.74 & 0.1913 & $\sim$200s \\
    SEVA~\cite{zhou2025stable} & 11.52 & 0.2231 & 0.5857 & 139.8 & 12.09 & 0.2315 & 11.30 & 0.2171 & $\sim$20s \\
    GLD~\cite{jang2026repurposing} & 12.03 & 0.1986 & 0.5631 & 162.1 & 13.32 & 0.2044 & 11.45 & 0.1900 & $\sim$42s \\
    Gen3R~\cite{huang2026gen3r} & 11.07 & 0.2056 & 0.6564 & 223.6 & 11.21 & 0.2072 & 10.96 & 0.2015 & $\sim$90s \\
    GEN3C~\cite{ren2025gen3c} & 12.86 & 0.2343 & 0.5796 & 207.4 & 14.43 & 0.2608 & 12.12 & 0.2117 & $\sim$1200s \\
    \textbf{Ours} & \textbf{13.10} & \textbf{0.2478} & \textbf{0.4895} & \textbf{139.4} & \textbf{15.37} & \textbf{0.2858} & \textbf{12.16} & \textbf{0.22} & \rev{$\sim$11s} \\
    \bottomrule
  \end{tabular}
\end{table}
\begin{figure}[t]
  \centering
  \includegraphics[width=0.92\linewidth]{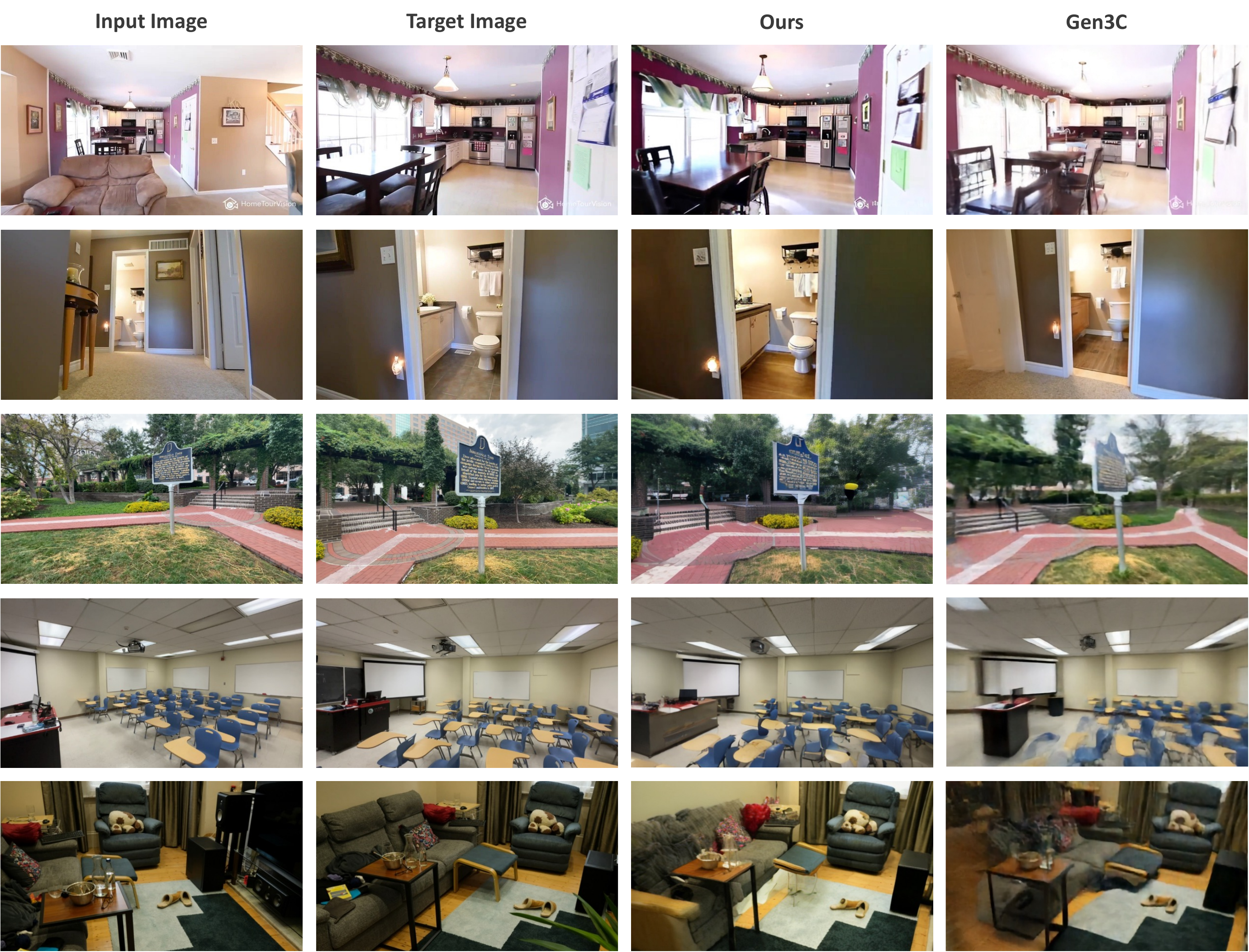}
  \caption{\textbf{Comparative qualitative results.} Single-view extrapolation against our strongest baseline, Gen3C~\citep{ren2025gen3c}. Each row shows the input view, ground-truth target, \methodname{}, and Gen3C.}
  \label{fig:comparative-figure}
\end{figure}



\noindent\textit{Quantitative Results.}
Tables~\ref{tab:results-re10k},~\ref{tab:results-dl3dv}, and~\ref{tab:results-mipnerf360} report quantitative results across the benchmarks. \methodname{} outperforms the baselines on every metric, demonstrating the effectiveness of our approach. In single-view extrapolation, baselines conditioned solely on Pl\"ucker rays are limited by pose-only scale ambiguity, while those using forward-warped RGB renders exploit metric-scale geometry for more consistent generations. Figure~\ref{fig:comparative-figure} compares qualitatively against our strongest baseline, Gen3C~\citep{ren2025gen3c}. Gen3C uses the forward-warped RGB render directly as the denoising target, yielding highly 3D-consistent outputs when warps are reliable; as camera motion grows, warp quality degrades and so do the generated images. This matches Gen3C's higher FID on DL3DV-10K and Mip-NeRF 360, which exhibit larger viewpoint changes than RE10K. We strongly encourage the readers to consult the supplementary for more examples. \rev{App.~\ref{app:split-ablation} ablates the generation--reconstruction split directly, and App.~\ref{app:depth-robustness} reports depth-robustness studies.}

\noindent\textbf{Two-View Conditioned Interpolation.}
\begin{table}[t]
  \caption{\textbf{Quantitative results on RE10K and DL3DV-10K.} Comparison with prior methods in the two-view interpolation setting.}
  \label{tab:results-interp}
  \centering
  \small
  \begin{tabular}{l ccc ccc}
    \toprule
    & \multicolumn{3}{c}{RE10K} & \multicolumn{3}{c}{DL3DV-10K} \\
    \cmidrule(lr){2-4} \cmidrule(lr){5-7}
    Method & PSNR $\uparrow$ & SSIM $\uparrow$ & LPIPS $\downarrow$ & PSNR $\uparrow$ & SSIM $\uparrow$ & LPIPS $\downarrow$ \\
    \midrule
    CameraCtrl~\cite{he2024cameractrl} & 12.87 & 0.4271 & 0.5022 & 15.60 & 0.3456 & 0.4705 \\
    ViewCrafter~\cite{yu2024viewcrafter} & 12.61 & 0.4550 & 0.4573 & 17.38 & 0.5214 & 0.3087 \\
    SEVA~\cite{zhou2025stable} & 16.15 & 0.5833 & 0.3477 & 21.38 & 0.6906 & 0.1845 \\
    LVSM~\cite{jin2024lvsm} & 16.88 & 0.5493 & 0.3983 & 21.31 & 0.6801 & 0.2223 \\
    GLD~\cite{jang2026repurposing} & 17.64 & 0.5707 & 0.2900 & 19.49 & 0.5575 & 0.2376 \\
    Gen3R~\cite{huang2026gen3r} & 12.17 & 0.4385 & 0.5697 & 13.99 & 0.3813 & 0.5030 \\
    GEN3C~\cite{ren2025gen3c} & 15.70 & 0.5632 & 0.3282 & 21.16 & 0.6869 & 0.2137 \\
    Ours & \textbf{19.80} & \textbf{0.7043} & \textbf{0.2205} & \rev{\textbf{21.99}} & \textbf{0.7009} & \textbf{0.1558} \\
    \bottomrule
  \end{tabular}
\end{table}
Next we evaluate \methodname{} in a predominantly reconstructive regime, where most of the information required to synthesize the target frames is already present in the two source views. We conduct this evaluation on RE10K and DL3DV-10K datasets. Table~\ref{tab:results-interp} reports reconstruction quality in the interpolation setting. \methodname{} outperforms all baselines on every reconstruction metric, confirming that our approach remains effective even when the task reduces to a nearly reconstruction-only (i.e., no generation) problem. \rev{App.~\ref{app:recon-baselines} extends this comparison to dedicated reconstruction methods, and App.~\ref{app:multiview} evaluates interpolation from four and six input views.}



\label{sec:ablations}
\begin{table}[t]
  \caption{\textbf{Ablation study on the effectiveness of individual components on RE10K and DL3DV.} Starting from a vanilla multi-view flow matching baseline, we progressively add the VAE skip connections + LoRA weights, and the reconstruction branch.}
  \label{tab:ablation-components}
  \centering
  \small
  \setlength{\tabcolsep}{4pt}
  \begin{tabular}{c l cccc cc}
    \toprule
    & & \multicolumn{4}{c}{Global} & \multicolumn{2}{c}{Observed} \\
    \cmidrule(lr){3-6} \cmidrule(lr){7-8}
    & Method & PSNR $\uparrow$ & SSIM $\uparrow$ & LPIPS $\downarrow$ & FID $\downarrow$ & PSNR $\uparrow$ & SSIM $\uparrow$ \\
    \midrule
    \multirow{3}{*}{\rotatebox[origin=c]{90}{RE10K}}
      & Generative Backbone & 15.86 & 0.5483 & 0.3402 & 37.82 & 19.20 & 0.6139 \\
      & + VAE Skip \& LoRA   & 15.86 & 0.5591 & 0.3394 & 35.27 & 19.23 & 0.6152 \\
      & + Reconstruction Branch & \rev{\textbf{17.05}} & \textbf{0.6071} & \textbf{0.3220} & \textbf{33.09} & \textbf{20.85} & \textbf{0.6738} \\
    \midrule
    \multirow{3}{*}{\rotatebox[origin=c]{90}{DL3DV}}
      & Generative Backbone & 18.62 & 0.5796 & 0.2556 & 39.94 & 19.96 & 0.6163 \\
      & + VAE Skip \& LoRA & 18.64 & 0.5828 & 0.2512 & 36.90 & 20.02 & 0.6215 \\
      & + Reconstruction Branch & \textbf{19.80} & \textbf{0.6195} & \textbf{0.2188} & \textbf{32.36} & \textbf{21.35} & \textbf{0.6605} \\
    \bottomrule
  \end{tabular}
\end{table}
\noindent\textbf{Ablating Individual Components.}
Table~\ref{tab:ablation-components} ablates the contribution of each architectural component on RE10K and DL3DV-10K. The VAE skip connections together with the decoder LoRA and the reconstruction branch play complementary roles. Adding the skip connections and LoRA weights primarily improves generative quality, as reflected in the lower FID scores on both datasets. The reconstruction branch, in contrast, sharpens reconstruction fidelity (PSNR, SSIM, LPIPS) without sacrificing generative quality.    
\section{Conclusion}
\label{sec:conclusion}

We presented \textbf{GenRec}, a multi-view flow matching model that treats generative novel-view synthesis as a structured composition of reconstruction and generation rather than a single distribution-matching problem. A per-pixel observation mask, derived from monocular depth and forward warping, governs the architecture, the loss, and the gradient flow: a flow matching backbone jointly denoises RGB and scene-coordinate latents to handle the distributional regime, while a pixel-space refinement stage with a skip- and LoRA-adapted decoder and a sparse 3D cross-attention branch restores high-frequency detail on observed pixels alone, with masked stop-gradients keeping its regression signal away from the generative prior. Across RealEstate10K, DL3DV-10K, and Mip-NeRF~360, in both single-view extrapolation and two-view interpolation, GenRec attains the best reconstruction fidelity in observed regions and matches or surpasses the strongest generative baselines on perceptual quality in unobserved ones, while running an order of magnitude faster than warp-as-target video baselines. 

\noindent\textbf{Limitations.} GenRec inherits the limitations of its monocular depth front-end, whose accuracy bounds the quality of the observation mask and warped conditioning, though monocular metric depth estimators continue to improve rapidly. Furthermore, computational constraints, compounded by our architecture's bi-modal nature, currently limit training to a limited number of views per scene. \rev{Apps.~\ref{app:failure-modes} and~\ref{app:depth-robustness} state our failure modes explicitly and quantify the depth dependence.}



\clearpage
\small
\bibliographystyle{plainnat}
\bibliography{references}
\raggedbottom


\appendix

\section*{Technical appendices and supplementary material}

\section{Qualitative Results}

\begin{figure}[H]
    \centering
    \includegraphics[width=\linewidth]{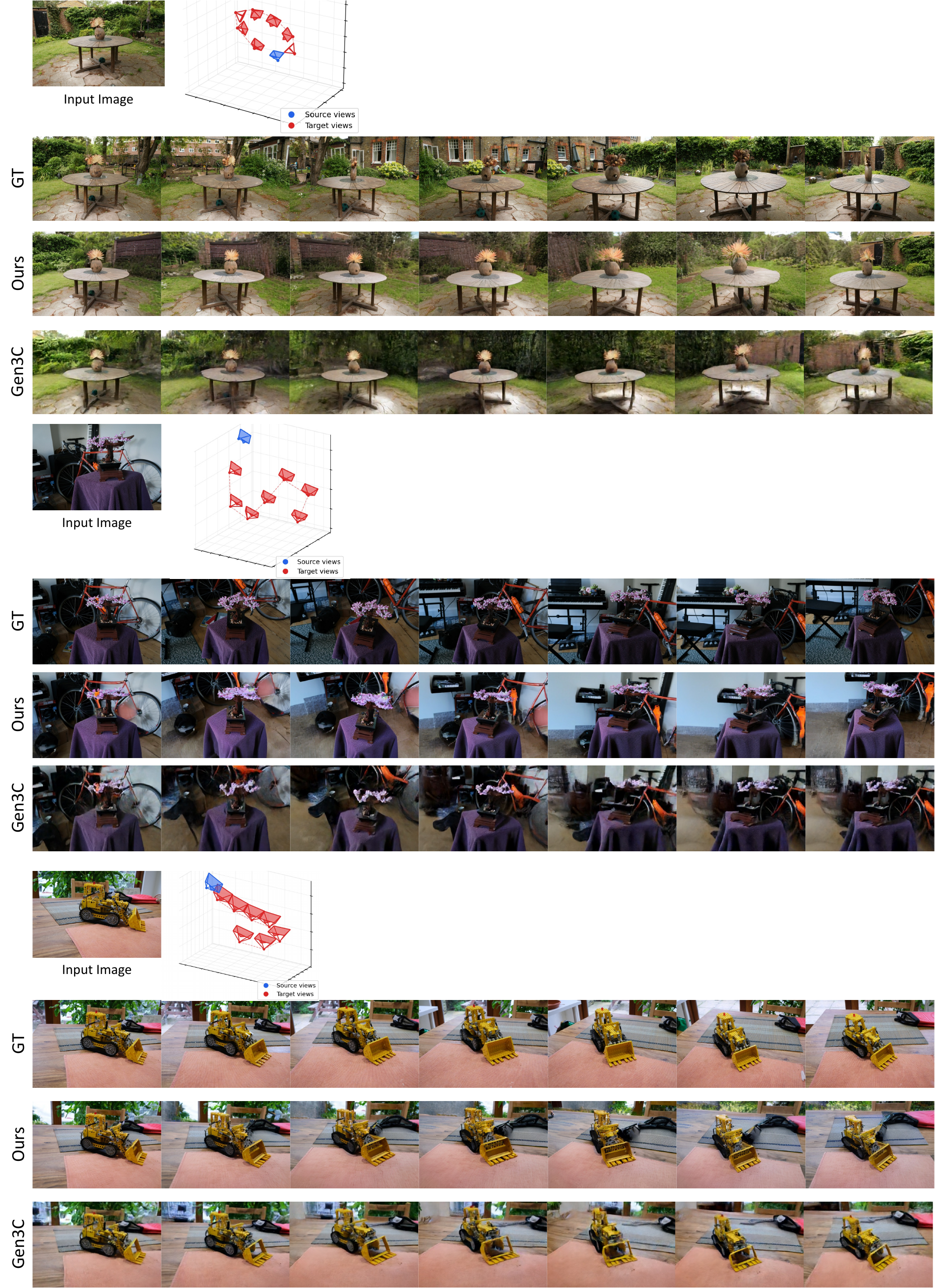}
    \caption{%
        \textbf{Qualitative comparison on Mip-NeRF~360}~\citep{barron2022mip}.
        Each row shows a held-out target view rendered from a single reference, compared against GEN3C. Mip-NeRF~360's unbounded outdoor scenes and wide-baseline trajectories stress geometric consistency in the unobserved background; our model preserves thin structures and produces sharper far-field detail than the baselines.
    }
    \label{fig:qual_mipnerf360}
\end{figure}

\begin{figure}[H]
    \centering
    \includegraphics[width=\linewidth]{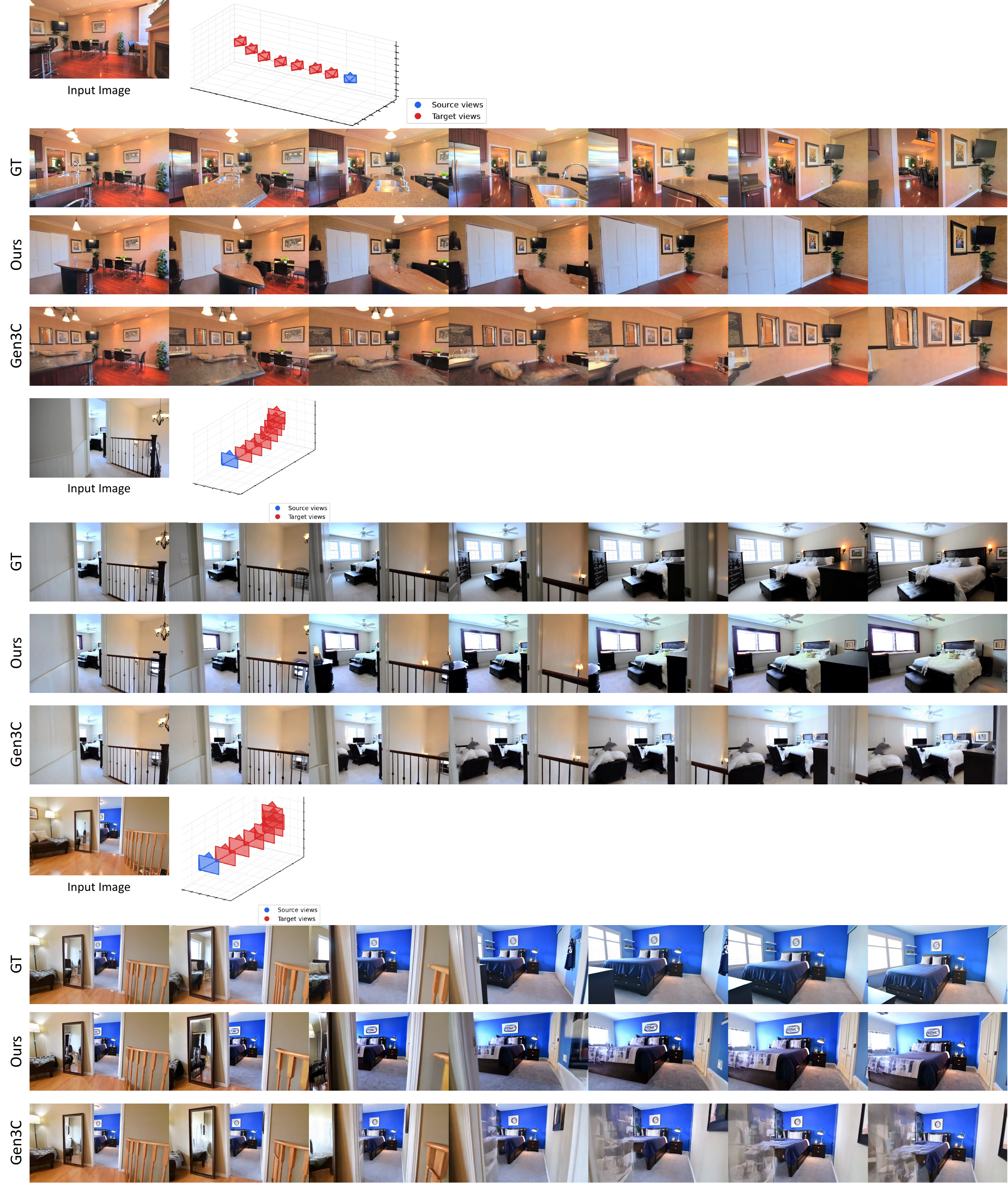}
    \caption{%
        \textbf{Qualitative comparison on RealEstate10K}~\citep{zhou2018stereo}.
        Indoor real-estate walkthroughs feature predominantly forward camera motion and narrow baselines. We highlight cases with disocclusions of structured surfaces (walls, doorways, furniture edges); our method recovers plausible texture in unobserved regions while baselines exhibit blur or ghosting along occlusion boundaries.
    }
    \label{fig:qual_re10k}
\end{figure}

\begin{figure}[H]
    \centering
    \includegraphics[width=\linewidth]{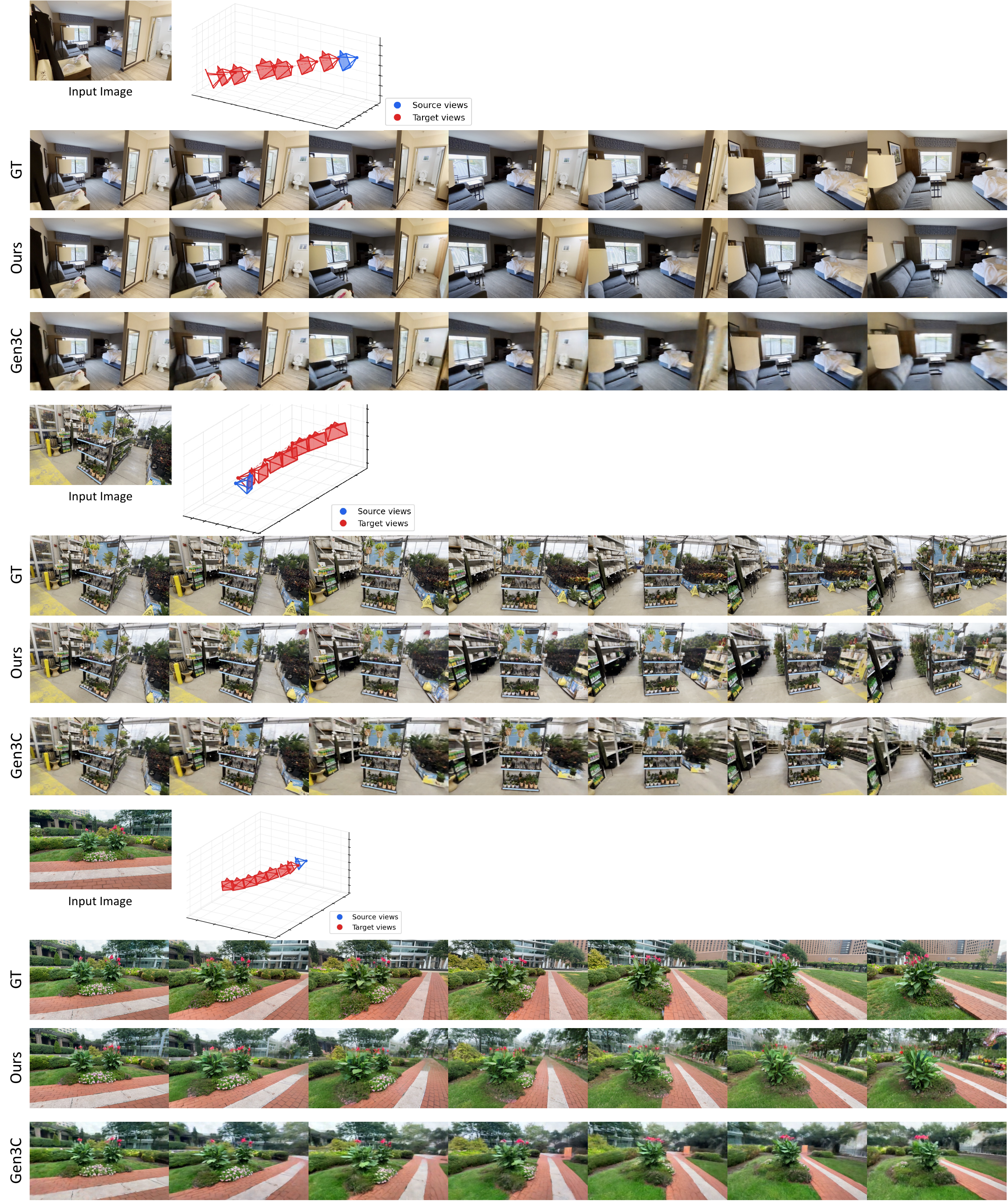}
    \caption{%
        \textbf{Qualitative comparison on DL3DV-10K}~\citep{ling2024dl3dv}.
        DL3DV-10K provides diverse, wide-baseline captures of indoor and outdoor
        scenes. The larger camera motions amplify the gap between methods that
        only inpaint warped pixels and methods that synthesize globally consistent
        geometry; our outputs remain sharp and view-consistent under aggressive
        target poses.
    }
    \label{fig:qual_dl3dv}
\end{figure}


\section{Additional Experiments and Ablation Studies}

\subsection{Effectiveness of the Generation--Reconstruction Split}
\label{app:split-ablation}
\begin{table}[H]
  \revcolor
  \caption{\textbf{Ablation of the generation--reconstruction split} on RealEstate10K, single-view extrapolation. \emph{Rec.}: dedicated pixel-space reconstruction branch; in (a) the branch is removed and the mask-gated regression loss is applied to the backbone directly. \emph{Gate}: mask-gated supervision, i.e., flow-matching supervision on all pixels with an additional regression supervision on observed ones. \emph{BP}: reconstruction gradients back-propagated into the backbone through the last $K{=}1$ sampling steps, following~\citet{clark2024directly}. Rows (b)--(d) share the same architecture, parameter count, and training procedure. Best results in \textbf{bold}.}
  \label{tab:supp-split-ablation}
  \centering
  \small
  \setlength{\tabcolsep}{3.5pt}
  \begin{tabular}{l ccc cccc cc cc}
    \toprule
    & & & & \multicolumn{4}{c}{Global} & \multicolumn{2}{c}{Observed} & \multicolumn{2}{c}{Unobserved} \\
    \cmidrule(lr){5-8} \cmidrule(lr){9-10} \cmidrule(lr){11-12}
    & Rec. & Gate & BP & PSNR $\uparrow$ & SSIM $\uparrow$ & LPIPS $\downarrow$ & FID $\downarrow$ & PSNR $\uparrow$ & SSIM $\uparrow$ & PSNR $\uparrow$ & SSIM $\uparrow$ \\
    \midrule
    (a) & $\times$ & \checkmark & -- & 16.51 & 0.5779 & 0.3459 & 36.84 & 19.88 & 0.6535 & 13.96 & 0.4667 \\
    (b) & \checkmark & $\times$ & $\times$ & 16.97 & 0.6020 & 0.3433 & 40.43 & 20.55 & 0.6724 & \textbf{14.28} & \textbf{0.4940} \\
    (c) & \checkmark & \checkmark & \checkmark & 16.99 & 0.6067 & 0.3429 & 34.92 & 20.77 & 0.6732 & 14.13 & 0.4738 \\
    \textbf{(d)} & \checkmark & \checkmark & $\times$ & \textbf{17.05} & \textbf{0.6071} & \textbf{0.3220} & \textbf{33.09} & \textbf{20.85} & \textbf{0.6738} & 14.18 & 0.4743 \\
    \bottomrule
  \end{tabular}
\end{table}

\rev{Table~\ref{tab:supp-split-ablation} isolates the effect of the generation--reconstruction split on RE10K in the single-view extrapolation setting. In row (a), the reconstruction branch is removed entirely, and the mask-gated supervision is instead applied to the generative backbone directly through a pixel-wise loss. In row (b), the branch is added but the gating is removed, so the branch is supervised on all pixels. In row (c), both the branch and the gating are present, but reconstruction gradients are back-propagated into the backbone through the last $K{=}1$ sampling steps, following~\citet{clark2024directly}. Row (d) is our full model: the branch is added, the gating is applied, and no reconstruction gradient reaches the backbone. Rows (b)--(d) share the same architecture, parameter count, and training procedure, differing only in whether supervision is gated by the observation mask and whether reconstruction gradients reach the backbone; row (a) removes the branch entirely as a reference point.

\noindent\textbf{Supervision --- rows (b) vs.\ (d).} These rows differ only in the mask gating. Removing the gating significantly degrades performance, with FID increasing by $7.34$. Although row (b) has more capacity than row (a), the lack of gating significantly damages the generative quality of the model.

\noindent\textbf{Gradient flow --- rows (c) vs.\ (d).} These rows differ only in whether reconstruction gradients reach the backbone. Allowing them through leaves row (c) behind the decoupled row (d) on every metric we measure, even though row (c) effectively has more trainable parameters through the additional finetuning of the backbone. The extra gradient flow from the reconstruction objective improves neither reconstruction fidelity nor generative quality.

\noindent\textbf{Architecture --- rows (a) vs.\ (d).} With gating applied in both, routing the regression signal through a separate, gradient-isolated pixel-space branch rather than applying it to the backbone directly yields significant improvements in both reconstruction fidelity (PSNR, SSIM) and generative quality (FID, LPIPS), indicating that the reconstruction-branch architecture is itself an important factor in the performance of the model.}

\subsection{Multi-View Interpolation Beyond the Training Regime}
\label{app:multiview}
\begin{table}[H]
  \revcolor
  \caption{\textbf{Multi-view interpolation on RE10K.} \methodname{} is trained with at most two input views, so the 4- and 6-view results are zero-shot (marked \emph{OOD}), whereas Gen3C's 3D cache ingests additional views natively. Best result per view count in \textbf{bold}.}
  \label{tab:supp-multiview}
  \centering
  \small
  \begin{tabular}{c l ccc}
    \toprule
    \#Views & Method & PSNR $\uparrow$ & SSIM $\uparrow$ & LPIPS $\downarrow$ \\
    \midrule
    \multirow{2}{*}{2}
      & Gen3C~\cite{ren2025gen3c} & 15.70 & 0.5632 & 0.3282 \\
      & \methodname{} & \textbf{19.80} & \textbf{0.7043} & \textbf{0.2205} \\
    \midrule
    \multirow{2}{*}{4}
      & Gen3C~\cite{ren2025gen3c} & 20.80 & 0.7596 & 0.1860 \\
      & \methodname{} (OOD) & \textbf{23.68} & \textbf{0.7988} & \textbf{0.1387} \\
    \midrule
    \multirow{2}{*}{6}
      & Gen3C~\cite{ren2025gen3c} & 20.86 & 0.7656 & 0.1793 \\
      & \methodname{} (OOD) & \textbf{24.07} & \textbf{0.8061} & \textbf{0.1342} \\
    \bottomrule
  \end{tabular}
\end{table}

\rev{The architecture places no cap on the number of source views: sources enter only through the warp and mask precomputation and through the key set of the sparse source-to-target attention, both linear in the number of source views. The two restrictions in our benchmarks are budgetary rather than architectural: the number of input views ($T_\text{in}\in\{1,2\}$) and the total number of views per training sample (App.~\ref{app:training}) bound the training sequence length, and neither is enforced at inference. Table~\ref{tab:supp-multiview} evaluates multi-view interpolation on RE10K against Gen3C, the strongest baseline in our comparisons and the closest to \methodname{} in conditioning. Since we train on at most two input views, the 4- and 6-view results are zero-shot, whereas Gen3C's 3D cache ingests additional views natively.

\methodname{} leads Gen3C on every metric at every view count, by $2.9$ to $4.1$~dB PSNR, despite never seeing more than two input views during training. The trend is more informative than the margins: between four and six views, \methodname{} gains $0.39$~dB while Gen3C gains $0.06$~dB and has effectively saturated. We read this through the paper's own lens. More source views raise the observed fraction of each target frame, exactly the regime the reconstruction branch exists to exploit, whereas a warp-as-target formulation stops benefiting once its cache already covers the target. The decomposition therefore scales with the available evidence rather than saturating with it, and does so zero-shot at four and six views.}

\subsection{Comparison with Reconstruction Methods}
\label{app:recon-baselines}
\begin{table}[H]
  \revcolor
  \caption{\textbf{Comparison with reconstruction methods} in the two-view interpolation setting on RE10K and DL3DV-10K. The baselines are specialised, deterministic reconstruction models; \methodname{} is generative. Best results in \textbf{bold}.}
  \label{tab:supp-recon-baselines}
  \centering
  \small
  \begin{tabular}{l ccc ccc}
    \toprule
    & \multicolumn{3}{c}{RE10K} & \multicolumn{3}{c}{DL3DV-10K} \\
    \cmidrule(lr){2-4} \cmidrule(lr){5-7}
    Method & PSNR $\uparrow$ & SSIM $\uparrow$ & LPIPS $\downarrow$ & PSNR $\uparrow$ & SSIM $\uparrow$ & LPIPS $\downarrow$ \\
    \midrule
    3DGS~\cite{kerbl20233d} & 14.95 & 0.543 & 0.421 & 19.76 & 0.681 & 0.248 \\
    Difix3D+~\cite{wu2025difix3d+} & 15.06 & 0.543 & 0.387 & 19.27 & 0.655 & 0.250 \\
    pixelSplat~\cite{charatan2024pixelsplat} & 16.96 & 0.614 & 0.392 & 19.29 & 0.625 & 0.328 \\
    MVSplat~\cite{chen2024mvsplat} & 16.64 & 0.592 & 0.383 & 17.53 & 0.535 & 0.366 \\
    DepthSplat~\cite{xu2025depthsplat} & \textbf{19.92} & \textbf{0.723} & 0.277 & 19.79 & 0.673 & 0.253 \\
    \methodname{} & 19.80 & 0.704 & \textbf{0.221} & \textbf{21.99} & \textbf{0.701} & \textbf{0.156} \\
    \bottomrule
  \end{tabular}
\end{table}

\rev{The two-view interpolation setting is predominantly reconstructive, so we additionally compare against dedicated reconstruction methods: 3DGS~\citep{kerbl20233d}, Difix3D+~\citep{wu2025difix3d+}, pixelSplat~\citep{charatan2024pixelsplat}, MVSplat~\citep{chen2024mvsplat}, and DepthSplat~\citep{xu2025depthsplat}, in Table~\ref{tab:supp-recon-baselines}. \methodname{} leads on all three metrics on DL3DV-10K ($+2.20$~dB PSNR over DepthSplat, with the best SSIM and LPIPS) and on LPIPS on both datasets. DepthSplat remains ahead on RE10K PSNR and SSIM, though \methodname{} trails by only $0.12$~dB. We attribute the remaining gap to a systematic effect: deterministic regressors trained under squared error tend toward the conditional mean, which PSNR and SSIM reward and LPIPS penalizes. These baselines are specialised reconstruction models, whereas \methodname{} is generative; it is therefore on par with them in the regime they are built for, while additionally extrapolating beyond the input views, which they cannot do.}

\subsection{3D Consistency}
\begin{table}[H]
    \centering
    \small
    \setlength{\tabcolsep}{6pt}
    \caption{\textbf{Camera pose accuracy and multi-view consistency on \textsc{RE10K} and \textsc{DL3DV-10K}.} ATE and RPE$_t$ are reported in the scale of the dataset camera poses; RPE$_r$ in degrees. Inference time is reported per scene. Lower is better for all metrics.}
    \begin{tabular}{lcccccc|r}
        \toprule
        &\multicolumn{3}{c}{\textsc{RE10K}}&\multicolumn{3}{c}{\textsc{DL3DV}}& \\
        \cmidrule(lr){2-4} \cmidrule(lr){5-7}
        Method&ATE $\downarrow$&RPE$_t$ $\downarrow$&RPE$_r$ $\downarrow$&ATE $\downarrow$&RPE$_t$ $\downarrow$&RPE$_r$ $\downarrow$&Inf. Time \\
        \midrule
        GLD~\cite{jang2026repurposing}&0.131&0.158&0.837&0.030&0.041&0.395&$\sim$42s \\
        ViewCrafter~\cite{yu2024viewcrafter}&0.134&0.148&1.446&0.045&0.060&0.535&$\sim$200s \\
        Gen3R~\cite{huang2026gen3r}&0.134&0.162&1.124&0.370&0.526&4.900&$\sim$90s \\
        Gen3C~\cite{ren2025gen3c}&\textbf{0.085}&\textbf{0.083}&\textbf{0.809}&0.037&0.052&0.430&$\sim$1200s \\
        \midrule
        Ours&0.087&0.106&1.048&\textbf{0.021}&\textbf{0.029}&\textbf{0.252}&\rev{$\sim$11s} \\
        \bottomrule
    \end{tabular}
    \label{tab:pose_consistency}
\end{table}
Beyond per-frame image quality, a generative novel view synthesis model should produce frames that are mutually consistent in 3D. We evaluate this property for \methodname{} in the single-view conditioned extrapolation setting on the two in-distribution datasets, RE10K and DL3DV-10K. We pass the generated frames through an off-the-shelf feed-forward 3D reconstruction model~\citep{wang2025vggt} to estimate their camera poses. Because VGGT recovers poses only up to a global similarity, we Sim(3)-align the estimates to the ground-truth target poses before computing Absolute Trajectory Error (ATE) and Relative Pose Error in rotation ($\text{RPE}_r$) and translation ($\text{RPE}_t$). The resulting values are therefore reported up to scale, but remain directly comparable across methods evaluated under the same protocol. Together, these metrics quantify how faithfully the generated frames adhere to the target pose conditioning.

Table~\ref{tab:pose_consistency} reveals a clear dataset-dependent pattern. On DL3DV-10K, \methodname{} achieves the best performance across all three metrics, with reductions of 29--36\% relative to the strongest baseline, GLD. On RE10K, Gen3C is the strongest method on every metric; \methodname{} performs comparably on ATE but underperforms on the relative-pose metrics, where it is also outperformed by GLD on rotation. We attribute this dataset-dependent ordering to a warp-quality dynamic: RE10K's narrow camera baselines yield reliable forward warps, which directly benefit Gen3C's warp-as-target formulation, whereas DL3DV-10K's larger viewpoint changes degrade warp quality and surface the advantages of \methodname{}'s observation-based design.

\subsection{Reconstruction Quality Across Observation Coverage}
\begin{figure}[H]
    \centering
    \begin{subfigure}[b]{0.48\linewidth}
        \includegraphics[width=\linewidth]{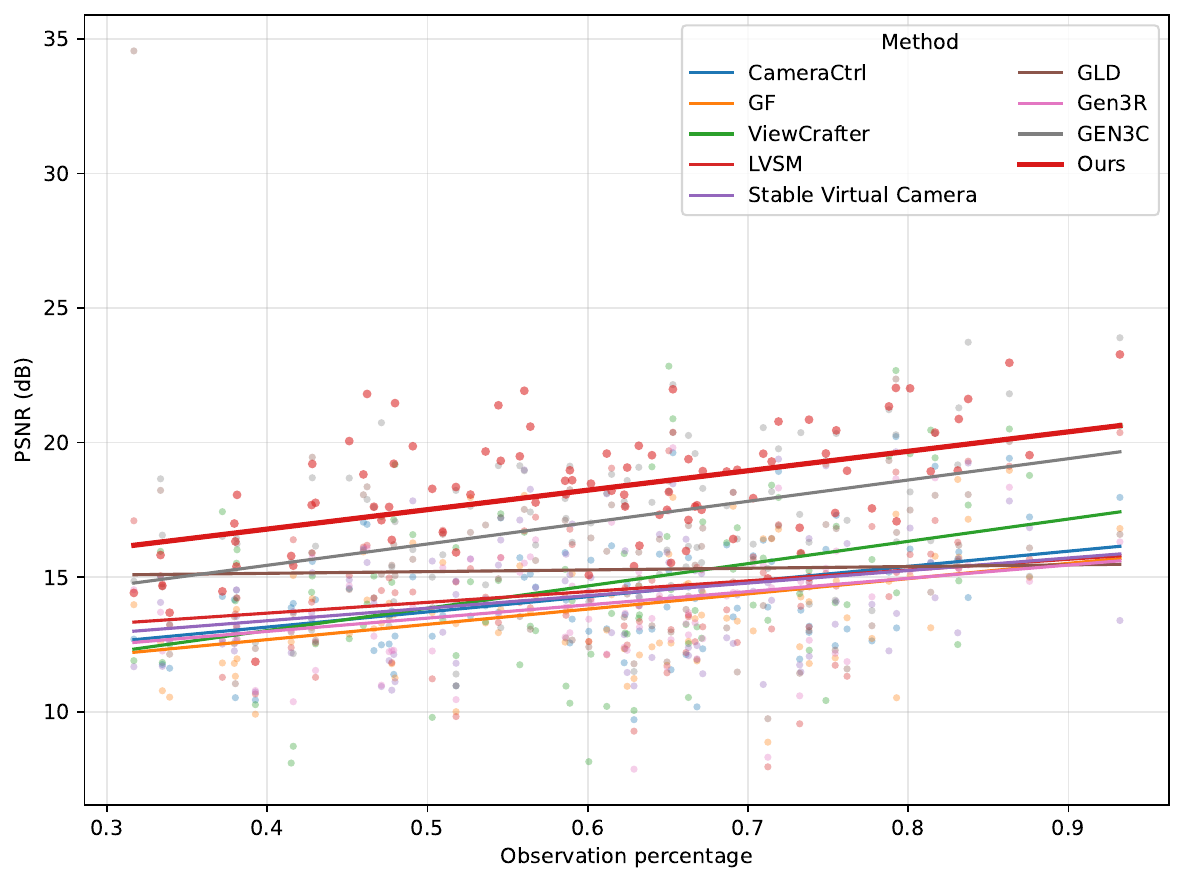}
        \caption{RealEstate10K.}
        \label{fig:sub1}
    \end{subfigure}
    \hfill
    \begin{subfigure}[b]{0.48\linewidth}
        \includegraphics[width=\linewidth]{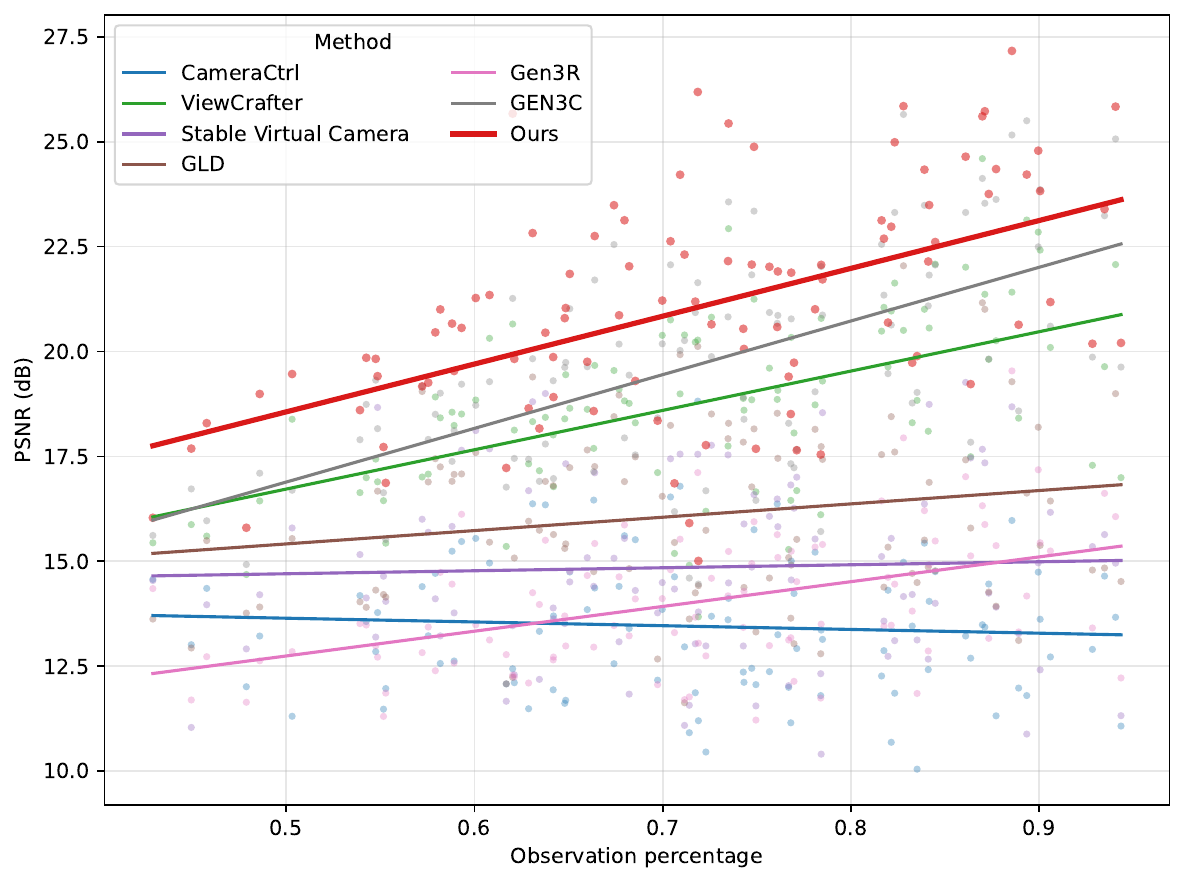}
        \caption{DL3DV-10K.}
        \label{fig:sub2}
    \end{subfigure}
    \caption{\textbf{PSNR as a function of per-scene observation percentage} on (a) RealEstate10K and (b) DL3DV-10K in the single-view conditioned setting. Each point is one test scene; lines are per-method linear fits. \methodname{} (red) sits above every baseline across the full range, from generation-heavy scenes on the left to near-reconstruction scenes on the right.}
    \label{fig:psnr_vs_obs}
\end{figure}

Figure~\ref{fig:psnr_vs_obs} stratifies single-view PSNR by per-scene observation percentage on RealEstate10K and DL3DV-10K, sweeping from generation-heavy scenes ($\sim$30\% observed) to near-reconstruction scenes ($>$90\% observed). \methodname{} dominates across the full range on both datasets: the regression line sits above every baseline at low observation rates, where most of the target view must be hallucinated, and remains the steepest at high observation rates, where the forward-warped point cloud is an accurate prior. This second regime is the more telling one. Methods such as GEN3C use the warped render directly as the denoising target and therefore inherit a strong reconstruction signal essentially for free in well-observed scenes. \methodname{} matches or exceeds them despite never tying its denoising target to the warped pixels, which we attribute to the reconstruction branch supplying the photometric fidelity that the latent generative path cannot. The result confirms that the generation--reconstruction split adapts gracefully to the actual observation budget of each scene.

\subsection{Effectiveness of Multi-Modal Generation}
\begin{table}[H]
  \caption{\textbf{Ablation on multi-modal generation.} We compare our joint RGB and scene-coordinate (SC) backbone against an RGB-only variant trained from the same initialization for 20K steps, evaluated in the single-view conditioned extrapolation setting.}
  \label{tab:ablation-multi-task}
  \centering
  \small
  \begin{tabular}{l cccc cccc}
    \toprule
    & \multicolumn{4}{c}{RE10K} & \multicolumn{4}{c}{DL3DV-10K} \\
    \cmidrule(lr){2-5} \cmidrule(lr){6-9}
    Method & PSNR $\uparrow$ & SSIM $\uparrow$ & LPIPS $\downarrow$ & FID $\downarrow$ & PSNR $\uparrow$ & SSIM $\uparrow$ & LPIPS $\downarrow$ & FID $\downarrow$ \\
    \midrule
    RGB & 15.09 & 0.5384 & 0.3602 & 38.22 & 18.05 & 0.5681 & 0.2631 & 38.53 \\
    RGB + SC (Ours) & \textbf{15.39} & \textbf{0.5457} & \textbf{0.3528} & \textbf{37.13} & \textbf{18.31} & \textbf{0.5749} & \textbf{0.2575} & \textbf{37.41} \\
    \bottomrule
  \end{tabular}
\end{table}
In Table~\ref{tab:ablation-multi-task} we isolate the effect of the joint RGB and scene-coordinate denoising introduced in Section~\ref{sec:backbone}. We train an RGB-only variant of the backbone that removes the scene-coordinate input channels, the scene-coordinate output head, and the zero-initialized cross-modal attention layers, leaving the cross-view attention and all other components unchanged. Both checkpoints are trained from the same Stable Diffusion 2.1~\cite{rombach2022high} initialization for 20K steps ($\sim$1/3 of full convergence) and evaluated under the single-view conditioned extrapolation protocol on RE10K and DL3DV-10K. 

Across both datasets, the multi-modal variant improves every metric, including FID and LPIPS, indicating that the joint denoising not only sharpens reconstruction in observed regions but also tightens the perceptual quality of the generated image. We attribute this to the cross-modal attention forcing the RGB branch to commit to a 3D-consistent interpretation of the scene at every denoising step, a regularizer that pure RGB diffusion lacks.

\subsection{Robustness to the Depth Front-End}
\label{app:depth-robustness}
\begin{table}[H]
  \revcolor
  \caption{\textbf{Depth source at inference} on RealEstate10K, single-view extrapolation. \methodname{} is trained on DA3 renders ($\dagger$); the estimator is swapped at test time with no retraining. Evaluated on the same RealEstate10K subset as Table~\ref{tab:supp-depth-robustness}, so values are not comparable to Table~\ref{tab:results-re10k}. Best results in \textbf{bold}.}
  \label{tab:supp-depth-source}
  \centering
  \small
  \setlength{\tabcolsep}{3.5pt}
  \begin{tabular}{l cccc cc cc}
    \toprule
    & \multicolumn{4}{c}{Global} & \multicolumn{2}{c}{Observed} & \multicolumn{2}{c}{Unobserved} \\
    \cmidrule(lr){2-5} \cmidrule(lr){6-7} \cmidrule(lr){8-9}
    Depth & PSNR $\uparrow$ & SSIM $\uparrow$ & LPIPS $\downarrow$ & FID $\downarrow$ & PSNR $\uparrow$ & SSIM $\uparrow$ & PSNR $\uparrow$ & SSIM $\uparrow$ \\
    \midrule
    DA3$^\dagger$~\cite{lin2025depth} & \textbf{17.35} & \textbf{0.6169} & \textbf{0.3206} & 63.69 & \textbf{22.34} & \textbf{0.6736} & \textbf{15.26} & \textbf{0.5242} \\
    MapAnything~\cite{keetha2025mapanything} & 17.23 & 0.6151 & 0.3236 & \textbf{63.66} & 22.08 & 0.6717 & 15.15 & 0.5192 \\
    VGGT~\cite{wang2025vggt} & 16.43 & 0.5734 & 0.3478 & 67.48 & 20.54 & 0.6092 & 14.79 & 0.4952 \\
    \bottomrule
  \end{tabular}
\end{table}

\rev{\methodname{} relies on a monocular depth front-end to build its observation mask and warped conditioning. We quantify this dependence with two studies.

\noindent\textbf{Swapping the depth estimator at inference.} Table~\ref{tab:supp-depth-source} replaces the depth front-end at inference only, measuring what is lost when depth no longer comes from the estimator the model was trained with (DA3~\citep{lin2025depth}). Swapping in MapAnything~\citep{keetha2025mapanything} costs $0.12$~dB PSNR and marginally improves FID, so the model is not tied to the estimator it was trained with; VGGT~\citep{wang2025vggt}, whose depth differs more, costs $0.92$~dB. Neither swap requires retraining, so \methodname{} also inherits future improvements in monocular depth for free: the front-end is a replaceable component, not a fixed dependency.

\begin{table}[H]
  \revcolor
  \caption{\textbf{Depth-robustness sweep} on RealEstate10K (25-scene subset, so clean values are not comparable to Tables~\ref{tab:results-re10k}--\ref{tab:results-mipnerf360}). At inference, the source depth is corrupted after the Umeyama alignment that fixes scale; every method builds its own conditioning from the corrupted depth, and all runs are scored against a single frozen observation mask. Each entry averages over the severity levels of the corresponding family: three levels for the affine corruptions, one for boundary, and three for noise. Best result per block in \textbf{bold}.}
  \label{tab:supp-depth-robustness}
  \centering
  \small
  \setlength{\tabcolsep}{3pt}
  \begin{tabular}{c l cccc cc cc}
    \toprule
    & & \multicolumn{4}{c}{Global} & \multicolumn{2}{c}{Observed} & \multicolumn{2}{c}{Unobserved} \\
    \cmidrule(lr){3-6} \cmidrule(lr){7-8} \cmidrule(lr){9-10}
    Corruption & Method & PSNR $\uparrow$ & SSIM $\uparrow$ & LPIPS $\downarrow$ & FID $\downarrow$ & PSNR $\uparrow$ & SSIM $\uparrow$ & PSNR $\uparrow$ & SSIM $\uparrow$ \\
    \midrule
    \multirow{3}{*}{Clean}
      & \methodname{} & \textbf{17.35} & \textbf{0.6196} & \textbf{0.3201} & \textbf{63.12} & \textbf{22.36} & \textbf{0.6759} & \textbf{15.25} & \textbf{0.5263} \\
      & Gen3C~\cite{ren2025gen3c} & 15.44 & 0.5613 & 0.3627 & 67.04 & 20.04 & 0.6037 & 13.81 & 0.4893 \\
      & ViewCrafter~\cite{yu2024viewcrafter} & 13.01 & 0.4749 & 0.4505 & 85.04 & 17.27 & 0.5222 & 11.43 & 0.3875 \\
    \midrule
    \multirow{3}{*}{Affine}
      & \methodname{} & \textbf{15.08} & 0.5122 & 0.4101 & \textbf{67.08} & \textbf{18.55} & 0.5232 & \textbf{14.09} & 0.4691 \\
      & Gen3C~\cite{ren2025gen3c} & 14.64 & \textbf{0.5240} & \textbf{0.4071} & 70.91 & 18.30 & \textbf{0.5404} & 13.56 & \textbf{0.4736} \\
      & ViewCrafter~\cite{yu2024viewcrafter} & 12.04 & 0.4217 & 0.4992 & 91.53 & 15.76 & 0.4491 & 10.92 & 0.3584 \\
    \midrule
    \multirow{3}{*}{Boundary}
      & \methodname{} & \textbf{17.23} & \textbf{0.6130} & \textbf{0.3251} & \textbf{65.87} & \textbf{22.09} & \textbf{0.6653} & \textbf{15.13} & \textbf{0.5169} \\
      & Gen3C~\cite{ren2025gen3c} & 15.53 & 0.5690 & 0.3620 & 66.95 & 19.73 & 0.5961 & 14.06 & 0.5042 \\
      & ViewCrafter~\cite{yu2024viewcrafter} & 12.83 & 0.4674 & 0.4589 & 89.49 & 17.02 & 0.5118 & 11.21 & 0.3776 \\
    \midrule
    \multirow{3}{*}{Noise}
      & \methodname{} & \textbf{16.53} & \textbf{0.5765} & \textbf{0.3666} & 76.48 & \textbf{20.55} & \textbf{0.5986} & \textbf{14.86} & \textbf{0.5077} \\
      & Gen3C~\cite{ren2025gen3c} & 15.25 & 0.5524 & 0.3793 & \textbf{67.96} & 19.32 & 0.5761 & 13.88 & 0.4902 \\
      & ViewCrafter~\cite{yu2024viewcrafter} & 11.66 & 0.3944 & 0.5202 & 103.82 & 15.12 & 0.4129 & 10.69 & 0.3403 \\
    \bottomrule
  \end{tabular}
\end{table}

\noindent\textbf{Stress-testing beyond real estimator error.} The second study corrupts the source depth at inference time, after the Umeyama alignment that fixes scale, using three families of corruption: an affine scale-and-shift in disparity, which models the residual ambiguity that survives alignment; boundary artifacts at depth discontinuities, which attack the observed/unobserved partition most directly; and spatially varying noise. Each family is evaluated at several severity levels, and Table~\ref{tab:supp-depth-robustness} reports the average over levels. Since every method builds its own conditioning from the corrupted depth, the corruption propagates to both the reconstruction branch and the backbone; all runs are scored against a single frozen observation mask, so the pixel sets remain fixed across comparisons. One caveat concerns Gen3C: its clean FID comes from the filter-disabled clean run, which is also the correct baseline for its noise rows, since those require the filter disabled (at the default setting, the renderer returns an empty mask at the strongest noise level). Measured against that run, Gen3C's noise loss is $0.53$~dB rather than the $0.20$~dB implied by the clean row.

Across the sweep, \methodname{} retains the PSNR lead in every corruption family and in every region, and the FID lead in all families except noise. The pattern of where ground is lost is more informative than the margins themselves. Boundary corruption, which attacks the observed/unobserved partition most directly, is the setting our design is built for, and it is also the one \methodname{} survives best. The affine family costs the most, exactly as the design predicts: affine rescaling deforms the warped evidence that the reconstruction branch consumes, so the path that benefits most from accurate depth is also the one most exposed to inaccurate depth. The same logic explains where the damage lands: in every family, the observed-region loss is two to five times the unobserved-region loss, i.e., the reconstruction path absorbs the depth error while the generative path remains comparatively insulated from it. Finally, the sweep perturbs the clean DA3 prediction rather than sensor depth, so it quantifies sensitivity to depth \emph{error} and should not be read as bounding absolute accuracy.}

\subsection{Failure Modes}
\label{app:failure-modes}
\rev{The studies above isolate concrete failure modes of the unified formulation and of \methodname{} itself, which we state explicitly here.

\noindent\textbf{Uniform supervision.} Row (b) of Table~\ref{tab:supp-split-ablation} is the unified-supervision configuration, and its failure is specific: fidelity on observed pixels improves while realism degrades (FID $40.43$, the weakest in the table), appearing as over-smoothed or mean-collapsed content in disocclusions.

\noindent\textbf{Depth-induced mask errors.} When depth is substantially wrong, the mask can label a genuinely unobserved pixel as observed, and the reconstruction branch then confidently reconstructs a mis-warped region rather than deferring to the generative prior. The depth-corruption sweep of Table~\ref{tab:supp-depth-robustness} confirms this mechanism: in every corruption family, the damage is two to five times larger in observed than in unobserved regions, exactly as the split predicts, while \methodname{} remains the strongest method on PSNR throughout.

\noindent\textbf{Very low coverage.} When a target view is almost entirely unobserved, the mask approaches zero and \methodname{} reduces to its backbone, so the split contributes little.

\noindent\textbf{Strong view-dependent effects.} On observed surfaces with strong view-dependent appearance, the source evidence available to the reconstruction branch is not the target appearance.}

\section{Preliminaries}
\label{app:preliminaries}

\subsection{Flow Matching}
Flow Matching~\citep{lipman2022flow} has emerged as a simpler and more efficient alternative to diffusion-based approaches for generative modeling. It is built around a time-dependent vector field $v : [0, 1] \times \mathbb{R}^d \to \mathbb{R}^d$, whose trajectories define a time-dependent diffeomorphism $\phi : [0, 1] \times \mathbb{R}^d \to \mathbb{R}^d$ called a \emph{flow}. For each initial point $x$, the curve $t \mapsto \phi_t(x)$ is the integral curve of $v_t$ starting at $x$, characterized by the ordinary differential equation:
\begin{equation}
    \frac{d}{dt} \phi_t(x) = v_t(\phi_t(x)), \quad \phi_0(x) = x.
\end{equation}
The flow $\phi_t$ induces a \emph{probability path} $p_t$ by pushing forward the initial distribution $p_0$ through the flow: $p_t = (\phi_t)_\# p_0$, where $p_0 = \mathcal{N}(0, I)$ is a standard Gaussian prior and $p_1 \approx q(x)$ is the target data distribution. The goal of Flow Matching is to learn a parametric vector field $v_\theta$ whose induced probability path transforms $p_0$ into $q$.

Following the optimal-transport conditional path of~\citet{lipman2022flow}, a sample at any intermediate time $t \in [0, 1]$ is obtained by linearly interpolating between a noise sample $x_0 \sim p_0$ and a data sample $x_1 \sim q$:
\begin{equation}
    x_t = (1 - t)\, x_0 + t\, x_1, \quad t \in [0,1].
\end{equation}
Differentiating the interpolant with respect to $t$ yields the conditional velocity field
\begin{equation*}
    u_t(x_t \mid x_0, x_1) = \frac{d}{dt}\bigl[(1-t)\, x_0 + t\, x_1\bigr] = x_1 - x_0,
\end{equation*}
which is constant along each interpolation trajectory. Because $x_t$ depends linearly on the endpoints, $x_t$ together with the velocity at time $t$ uniquely determines $x_1$. Rearranging the interpolant yields a closed-form estimate of the clean sample at any $t \in [0, 1)$:
\begin{equation}
    \hat{x}_1 = x_t + (1 - t)\, v_\theta(x_t, t),
    \label{eq:x1_pred}
\end{equation}
so the learned velocity network $v_\theta$ implicitly acts as an $x_1$-predictor at every timestep. A single forward pass at arbitrary $t$ therefore yields both a denoising direction and an estimate of the target, without the noise-level-dependent reweighting required by $\epsilon$-prediction in diffusion models~\citep{ho2020denoising}.

The network $v_\theta$ is trained with the Conditional Flow Matching (CFM) objective~\citep{lipman2022flow}, which regresses onto the per-sample target velocity:
\begin{equation}
    \mathcal{L}_{\mathrm{CFM}}(\theta) = \mathbb{E}_{t,\, x_0,\, x_1}\,\bigl\| v_\theta(x_t, t) - (x_1 - x_0) \bigr\|^2,
    \label{eq:cfm_loss}
\end{equation}
where $x_0 \sim \mathcal{N}(0, I)$, $x_1 \sim q(x)$, and $x_t = (1-t)\, x_0 + t\, x_1$. We follow~\citet{esser2024scaling} and draw $t$ from a logit-normal distribution, $t = \sigma(u)$ for training. Because both the interpolated sample and its target velocity are available in closed form, this loss admits unbiased stochastic estimates and is \emph{simulation-free}, meaning training never requires integrating the ODE. At inference, a data sample is generated by drawing $x_0 \sim \mathcal{N}(0, I)$ and integrating the learned vector field from $t = 0$ to $t = 1$. Discretizing the interval into $N$ steps with $\Delta t = 1/N$, the simplest Euler update advances the state from $t_k$ to $t_{k+1} = t_k + \Delta t$ as
\begin{equation}
    x_{t_{k+1}} = x_{t_k} + \Delta t \cdot v_\theta(x_{t_k}, t_k),
    \label{eq:euler_step}
\end{equation}
which can be replaced with any higher-order ODE solver (e.g. Heun or Runge--Kutta) for improved accuracy at the same number of function evaluations. We utilize the Euler update in our work as shown in Equation \ref{eq:euler_step}.

\subsection{Diff2Flow~\citep{schusterbauer2025diff2flow}}

While Flow Matching offers attractive properties such as straighter sampling trajectories and simulation-free training, the most widely available early foundation generative models (e.g., Stable Diffusion~\citep{rombach2022high}) are trained as denoising diffusion models~\citep{ho2020denoising,song2020score} rather than flow models. Diff2Flow bridges this gap by aligning the two paradigms through coupled reparameterizations.

A diffusion model corrupts a data sample $x_0^{\mathrm{DM}} \sim q(x)$ with Gaussian noise $x_T^{\mathrm{DM}} \sim \mathcal{N}(0, I)$ along discrete timesteps $t_{\mathrm{DM}} \in \mathbb{Z}_{\geq 0} \cap [0, T]$. Under a variance-preserving schedule, the noisy interpolant is
\begin{equation}
    x_{t_{\mathrm{DM}}}^{\mathrm{DM}} = \alpha_{t_{\mathrm{DM}}}\, x_0^{\mathrm{DM}} + \sigma_{t_{\mathrm{DM}}}\, x_T^{\mathrm{DM}}, \qquad \alpha_{t_{\mathrm{DM}}}^2 + \sigma_{t_{\mathrm{DM}}}^2 = 1,
    \label{eq:dm_interpolant}
\end{equation}
with $\alpha_{t_{\mathrm{DM}}}$ monotonically decreasing and $\sigma_{t_{\mathrm{DM}}}$ monotonically increasing in $t_{\mathrm{DM}}$. The network is trained either to regress the noise $\epsilon$ or, in the $\boldsymbol{v}$-parameterization~\citep{salimans2022progressive}, the target
\begin{equation}
    v_{t_{\mathrm{DM}}}^{\mathrm{DM}} = \alpha_{t_{\mathrm{DM}}}\, x_T^{\mathrm{DM}} - \sigma_{t_{\mathrm{DM}}}\, x_0^{\mathrm{DM}},
    \label{eq:v_target}
\end{equation}
the parameterization used by Stable Diffusion 2.1, on which we focus throughout. Note the convention mismatch with Flow Matching: the diffusion timestep treats $x_0^{\mathrm{DM}}$ as data and $x_T^{\mathrm{DM}}$ as noise, opposite to the FM convention introduced above. The two are reconciled by the boundary identification $x_0^{\mathrm{DM}} \equiv x_1$ and $x_T^{\mathrm{DM}} \equiv x_0$.

To align the trajectories, Diff2Flow constructs invertible reparameterizations $f_t : [0, T] \to [0, 1]$ on time and $f_x : \mathbb{R}^d \to \mathbb{R}^d$ on the interpolant, chosen such that the diffusion trajectory in Eq.~\eqref{eq:dm_interpolant} coincides with the FM trajectory $x_t = (1-t)\, x_0 + t\, x_1$ under the above boundary identification. Equating the two expressions and solving for $t$ in terms of the diffusion coefficients yields
\begin{equation}
    f_t(t_{\mathrm{DM}}) = \frac{\alpha_{t_{\mathrm{DM}}}}{\alpha_{t_{\mathrm{DM}}} + \sigma_{t_{\mathrm{DM}}}}, \qquad
    f_x\bigl(x_{t_{\mathrm{DM}}}^{\mathrm{DM}}\bigr) = \frac{1}{\alpha_{t_{\mathrm{DM}}} + \sigma_{t_{\mathrm{DM}}}}\, x_{t_{\mathrm{DM}}}^{\mathrm{DM}}.
    \label{eq:diff2flow_maps}
\end{equation}
Under both variance-preserving and variance-exploding schedules $f_t$ is strictly monotonic and hence invertible. The discrete coefficients $\{\alpha_{t_{\mathrm{DM}}}, \sigma_{t_{\mathrm{DM}}}\}$ are linearly interpolated between their nearest integer neighbors so that $f_t$ and $f_t^{-1}$ are well-defined on the entire continuous interval; this extension is justified empirically by the observation that diffusion networks trained on integer timesteps produce coherent outputs at fractional ones~\citep{schusterbauer2025diff2flow}. Given an FM timestep $t \in [0, 1]$ and interpolant $x_t$, the diffusion-side quantities are recovered as
\begin{equation}
    t_{\mathrm{DM}} = f_t^{-1}(t), \qquad
    x_{t_{\mathrm{DM}}}^{\mathrm{DM}} = f_x^{-1}(x_t) = \bigl(\alpha_{t_{\mathrm{DM}}} + \sigma_{t_{\mathrm{DM}}}\bigr)\, x_t.
    \label{eq:diff2flow_inverse}
\end{equation}

Rather than fine-tuning the diffusion network to directly emit FM velocities, which forces it to abandon its existing parameterization and degrades performance under parameter-efficient adaptation~\citep{schusterbauer2025diff2flow}, Diff2Flow keeps the network in its native parameterization and analytically converts its output. Solving the linear system formed by Eq.~\eqref{eq:dm_interpolant} and Eq.~\eqref{eq:v_target} for the endpoints gives
\begin{equation}
    \hat{x}_0^{\mathrm{DM}} = \alpha_{t_{\mathrm{DM}}}\, x_{t_{\mathrm{DM}}}^{\mathrm{DM}} - \sigma_{t_{\mathrm{DM}}}\, v_\theta^{\mathrm{DM}}\bigl(x_{t_{\mathrm{DM}}}^{\mathrm{DM}}, t_{\mathrm{DM}}\bigr),
    \qquad
    \hat{x}_T^{\mathrm{DM}} = \sigma_{t_{\mathrm{DM}}}\, x_{t_{\mathrm{DM}}}^{\mathrm{DM}} + \alpha_{t_{\mathrm{DM}}}\, v_\theta^{\mathrm{DM}}\bigl(x_{t_{\mathrm{DM}}}^{\mathrm{DM}}, t_{\mathrm{DM}}\bigr).
    \label{eq:dm_endpoints}
\end{equation}
Substituting the boundary identification into the FM target $x_1 - x_0$ then yields a closed-form FM velocity expressed entirely through the diffusion network's $\boldsymbol{v}$-prediction:
\begin{equation}
    v_\theta(x_t, t)
    \;=\; \hat{x}_0^{\mathrm{DM}} - \hat{x}_T^{\mathrm{DM}}
    \;=\; (\alpha_{t_{\mathrm{DM}}} - \sigma_{t_{\mathrm{DM}}})\, x_{t_{\mathrm{DM}}}^{\mathrm{DM}}
    \;-\; (\alpha_{t_{\mathrm{DM}}} + \sigma_{t_{\mathrm{DM}}})\, v_\theta^{\mathrm{DM}}\bigl(x_{t_{\mathrm{DM}}}^{\mathrm{DM}}, t_{\mathrm{DM}}\bigr),
    \label{eq:diff2flow_velocity}
\end{equation}
with $t_{\mathrm{DM}} = f_t^{-1}(t)$ and $x_{t_{\mathrm{DM}}}^{\mathrm{DM}} = f_x^{-1}(x_t)$ given by Eqs.~\eqref{eq:diff2flow_maps}--\eqref{eq:diff2flow_inverse}. An analogous derivation holds for $\epsilon$-parameterized priors.

Training proceeds with the standard CFM loss in Eq.~\eqref{eq:cfm_loss}, where the velocity prediction is evaluated through Eq.~\eqref{eq:diff2flow_velocity}. Sampling uses the Euler update of Eq.~\eqref{eq:euler_step} on the FM trajectory: at each step, the current state $x_{t_k}$ and timestep $t_k$ are mapped to the diffusion side via $f_t^{-1}$ and $f_x^{-1}$, the network is queried in its native parameterization, and its $\boldsymbol{v}$-prediction is converted back to an FM velocity through Eq.~\eqref{eq:diff2flow_velocity}.

\paragraph{Inference Protocol.}

At test time we sample with $50$ solver steps using the EMA weights accumulated during training. We employ classifier-free guidance with a scale of $1.5$.

\subsection{Implementation Details}
\label{app:impl}

\paragraph{Generative Backbone.}
We initialize the multi-view generative backbone from the Stable Diffusion 2.1 (SD-2.1) checkpoint~\citep{rombach2022high} and zero-initialize the cross-modal attention layers. The number of attention heads and head dimensions are reused from SD-2.1 and is symmetric for both cross-view and cross-modal attention. We reuse the pretrained SD-2.1 VAE for the image modality and adopt the scene-coordinate VAE from SpatialGen~\citep{fang2025spatialgen} for the scene-coordinate modality; both VAEs remain frozen throughout training. Pl\"ucker ray conditioning is embedded via the SpatialGen ray encoder.

\paragraph{VAE Skip Adapter and LoRA.}
The mask-conditioned skip adapter (Section~\ref{sec:backbone}) uses a bias-free $1{\times}1$ convolution per skip, initialized with standard deviation $10^{-5}$ so that an untrained adapter reproduces the pretrained decoder within numerical noise. We adapt the decoder with low-rank residuals via PEFT~\citep{peft}, targeting every convolution and attention projection together with the four skip convolutions: rank $r=8$, scaling $\alpha=r$, Gaussian initialization, biases untouched.

\paragraph{Reconstruction Branch.}
The branch operates at an internal width of $D{=}64$ with four FiLM-conditioned residual blocks interleaved with sparse geometric attention layers (4 heads, head dimension $16$). The attention layers between the  The query and source streams are projected to $D$ via $3{\times}3$ and $1{\times}1$ convolutions respectively. Sparsity in both the geometric cross-attention and self-attention is enforced through precomputed $k$-nearest-neighbour index maps with $k{=}8$.

\subsection{Training Details}
\label{app:training}
Our pipeline follows two stages: we first train the generative backbone together with the VAE skip connections and the decoder LoRA weights, then freeze it and train the reconstruction branch with iterative denoising supervision.

\paragraph{Data.}
We train our method on a mixture of RealEstate10K~\citep{zhou2018stereo}, DL3DV-10K~\citep{ling2024dl3dv} and SpatialVID~\citep{wang2025spatialvid} with sampling weights $0.5$, $0.2$ and $0.3$ respectively, all loaded at a longest-side resolution of $616$ pixels with dataset-specific frame strides. From SpatialVID we filter out dynamic scenes by thresholding the depth-projection error between frames computed with DA3 depth, retaining only scenes whose static-geometry assumption holds. These datasets together form a collection of around $160$K indoor and outdoor multi-view consistent scenes. For every scene we sample $T=8$ consecutive frames at the dataset-specific stride; the number of clean input views $T_\text{in}$ is drawn from $\{1, 2\}$ with probabilities $0.8$ and $0.2$, placed at the first $T_\text{in}$ positions of the sampled window, and the remaining $T_\text{out} = T - T_\text{in}$ views are noised. Metric depth is computed by Depth Anything 3 (DA3)~\citep{lin2025depth} at $504$ resolution, matching DA3's base training resolution. We use Umeyama alignment between the predicted camera poses from DA3 and the ground-truth poses to scale the depth estimation for training.

\paragraph{Multi-view Generation Backbone.}
We train the generative backbone for $20$ epochs ($\sim$72K optimizer steps) with an effective batch size of $16$ using 8-bit AdamW, learning rate $10^{-5}$, no weight decay, and a global gradient-norm clip of $1.0$. The learning rate is linearly warmed up over $500$ steps and decayed with a cosine schedule to a minimum of $10^{-6}$. To enable classifier-free guidance at inference, we drop the geometric conditioning (warped renders and observation mask) independently with probability $0.1$ during training. Training is performed in \texttt{bfloat16} mixed precision with gradient checkpointing enabled throughout, distributed across 16 GH200 GPUs via PyTorch DDP for a total wall-clock time of approximately three days. We maintain an exponential-moving-average copy of the weights with decay $0.9999$.

\paragraph{Reconstruction Branch.}
The branch is trained in the same optimization loop as the frozen generative backbone but with fully decoupled gradients: gradients influence only the residual predictor. To match the distribution of samples encountered at inference, the detached $\hat{x}^I_1$ is produced by an unrolled denoising rollout of $50$ no-gradient solver steps, terminated at a randomly chosen step whose corresponding diffusion timestep, recovered from the flow time $t$ via the Diff2Flow mapping $f_t^{-1}$ on SD2.1's $1000$-step schedule, falls below $200$. This exposes the branch to realistic, near-final samples rather than the single-step estimates available during diffusion training. The branch parameters form a separate optimizer group with learning rate $5{\times}10^{-5}$, weight decay $10^{-4}$, and a $1000$-step linear warm-up; all other settings (8-bit AdamW, gradient-norm clip $1.0$, \texttt{bfloat16} mixed precision) match those of the backbone. The branch converges within $10$K steps.

\paragraph{Loss Weights.}
For the geometry-aware decoder adapter (Eq.~\ref{eq:decoder_loss}), we set $\lambda_1 = 0.1$ and $\lambda_{\text{p}} = 0.3$. For the pixel-space reconstruction branch (Eq.~\ref{eq:recon_loss}), we set $\lambda_1 = 0.1$, $\lambda_{\text{p}} = 0.5$, and $\lambda_{\Delta} = 0.05$.

\section{Societal Impacts}
\label{sec:societal}

GenRec synthesizes novel views of real-world scenes from one or two posed
images, with applications in robotic navigation under sparse observation,
mixed-reality and accessibility. The order-of-magnitude inference speedup over warp-as-target video baselines makes such uses practical on commodity hardware.

As with any photorealistic generative model, outputs could in principle be misused to fabricate imagery, and the learned prior reflects the distribution of its training data. We train on RealEstate10K, DL3DV-10K, and SpatialVID, which are publicly released, mostly widely adopted benchmarks in the 3D vision community; the resulting biases are those already characterized in the literature surrounding these datasets, and the privacy considerations are governed by the curation policies under which they were released. We do not introduce new data collection.

A property of GenRec's design that is directly relevant here is that the
observation mask $O_i$ separates pixels recovered by reconstruction from
pixels produced by the generative prior, and this separation is available at
inference at no additional cost. The mask functions as a built-in provenance
signal: it identifies, per pixel, whether the model is reporting evidence or
imagining a plausible completion. We encourage downstream systems to surface
this signal (through overlays, metadata, or content-credential standards
such as C2PA) rather than discarding it once the final image is composited.



\end{document}